\documentclass{article}

\usepackage[toc,page]{appendix}
\usepackage[utf8]{inputenc}
\usepackage{float}
\usepackage{pdfpages}
\usepackage{subcaption}
\usepackage{graphicx}
\usepackage{subcaption}
\usepackage{hyperref}
\usepackage[T1]{fontenc}
\usepackage[utf8]{inputenc}
\usepackage[english]{babel}
\usepackage{graphicx}
\usepackage{tikz} 
\usepackage{tocloft}
\graphicspath{{./figs/}}
\usepackage{setspace}
\usepackage[super]{nth}
\usepackage{amsmath}

\usepackage{float}
\usepackage{arxiv}

\usepackage[utf8]{inputenc} % allow utf-8 input
\usepackage[T1]{fontenc}    % use 8-bit T1 fonts
\usepackage{hyperref}       % hyperlinks
\usepackage{url}            % simple URL typesetting
\usepackage{booktabs}       % professional-quality tables
\usepackage{amsfonts}       % blackboard math symbols
\usepackage{nicefrac}       % compact symbols for 1/2, etc.
\usepackage{microtype}      % microtypography
\usepackage{lipsum}		% Can be removed after putting your text content
\usepackage{graphicx}
\usepackage[numbers]{natbib}
\usepackage{doi}

\title{A Comparative Study of NeuralODE and Universal ODE Approaches to Solving Chandrasekhar's White Dwarf Equation}

\author{ \href{https://orcid.org/0000-0000-0000-0000}{\includegraphics[scale=0.06]{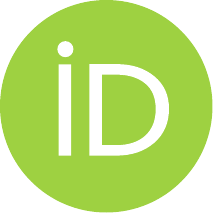}\hspace{1mm}Raymundo Vazquez Martinez}\thanks{Use footnote for providing further
		information about author (webpage, alternative
		address)---\emph{not} for acknowledging funding agencies.} \\
	National Autonomous University of Mexico\\
	\texttt{raymundo.vm@ciencias.unam.mx} \\
	\And
	\href{https://orcid.org/0000-0000-0000-0000}{\includegraphics[scale=0.06]{orcid.pdf}\hspace{1mm}Raj Abhijit Dandekar} \\
	Vizuara AI Labs\\
        Massachusetts Institute of Technology (prior)\\
	\texttt{raj@vizuara.com} \\
 \And
	\href{https://orcid.org/0000-0000-0000-0000}{\includegraphics[scale=0.06]{orcid.pdf}\hspace{1mm}Rajat Dandekar} \\
	Vizuara AI Labs\\
       Purdue University (prior)\\
	\texttt{rajatdandekar@vizuara.com} \\
 \And
	\href{https://orcid.org/0000-0000-0000-0000}{\includegraphics[scale=0.06]{orcid.pdf}\hspace{1mm}Sreedath Panat} \\
	Vizuara AI Labs\\
         Massachusetts Institute of Technology (prior)\\
	\texttt{sreedath@vizuara.com} \\
}
\renewcommand{\shorttitle}{\textit{arXiv} Template}

\hypersetup{
pdftitle={A template for the arxiv style},
pdfsubject={q-bio.NC, q-bio.QM},
pdfauthor={David S.~Hippocampus, Elias D.~Striatum},
pdfkeywords={Chadrasekhar's white dwarf, SciML, UDE},
}

\begin{document}
\maketitle

\begin{abstract}
	In this study, we apply two pillars of Scientific Machine Learning: Neural Ordinary Differential Equations (Neural ODEs) and Universal Differential Equations (UDEs) to the Chandrasekhar White Dwarf Equation (CWDE). The CWDE  is fundamental for understanding the life cycle of a star, and describes the relationship between the density of the white dwarf and it's distance from the center. Despite the rise in Scientific Machine Learning frameworks, very less attention has been paid to the systematic applications of the above SciML pillars on astronomy based ODEs. Through robust modeling in the Julia programming language, we show that both Neural ODEs and UDEs can be used effectively for both prediction as well as forecasting of the CWDE. More importantly, we introduce the "forecasting breakdown point" - the time at which forecasting fails for both Neural ODEs and UDEs. Through a robust hyperparameter optimization testing, we provide insights on the neural network architecture, activation functions and optimizers which provide the best results. This study provides opens a door to investigate the applicability of Scientific Machine Learning frameworks in forecasting tasks for a wide range of scientific domains. 

\end{abstract}

% keywords can be removed
\section{Introduction}
Scientific Machine Learning (Scientific ML) is a growing field with a wide range of applications in various fields such as epidemiology, gene expression, optics, circuit modeling, quantum circuits and fluid mechanics \cite{scimlappl1, scimlappl2, scimlappl3, scimlappl4, scimlappl5, scimlappl6, scimlappl7, scimlappl8, scimlappl9, scimlappl10, scimlappl11, scimlappl12, scimlappl13, scimlappl14}. This field of Scientific ML leverages the interpretibility of scientific structures like ODEs/PDEs along with the expressivity of neural networks. Broadly, the rise of Scientific Machine Learning can be attributed to three popular methodologies:

\begin{itemize}
    \item Neural Ordinary Differential Equations: The entire forward pass of an ODE/PDE is replaced with neural networks. We perform backpropagation through the neural network augmented ODE/PDE. In doing so, we find the optimal values of the neural network parameters. \cite{node1, node2, node3, node4}
    \item Universal Differential Equations (UDEs): In contrast to Neural ODEs, only certain terms of the ODE/PDEs are replaced with neural networks. We then discover these terms by optimizing the neural network parameters. Universal Differential Equations can be used to correct existing underlying ODEs/PDEs as well as to discover new, missing physics. \cite{ude1, ude2, ude3, ude4}
    \item Physics Informed Neural Networks (PINNs): PINNs are predominantly used as an alternative to traditional ODE/PDE solvers to solve an entire ODE/PDE. We replace the variable with a neural network and the loss function is determined by the ODE/PDE solution and the boundary conditions. When we minimize the loss function, we automatically find the optimium solution to the ODE/PDE. \cite{pinn1, pinn2, pinn3, pinn4}
\end{itemize}

Despite the advances of Scientific ML in various fields, there is a lack of applying Scientific ML methods in the field of astronomy. Although there are a few studies aimed at applying Neural ODEs on astronomy problems \cite{astrosciml1, astrosciml2, astrosciml3}, there is no study investigating the application of Universal Differential Equations (UDEs) on astronomy problems. 

In particular, the following questions are still unanswered:

\begin{itemize}
    \item In the spirit of UDEs, can we replace certain terms of an astronomical ODE system with neural networks and recover them?
    \item How does the Neural ODE prediction compare with the UDE prediction?
    \item Can we do forecasting on the system of ODEs with Neural ODEs and UDEs?
    \item Are UDEs better at forecasting than Neural ODEs?
\end{itemize}

We aim to answer these questions by looking at a foundational ODE in astronomy: the Chandrasekhar White Dwarf equation (CWDE) \cite{cwde1, cwde2}. The CWDE describes the relationship between the density of the white dwarf and it's distance from the center \cite{cwde1, cwde2}. This equation is fundamental for understanding the life cycle of a star. Using this equation, we can potentially predict when the star will collapse and transform into a supernova.

We use the advanced Scientific Machine Learning libraries provided by the Julia Programming Language \cite{Julia1, Julia2, Julia3, Julia4}. Through a robust hyperparameter optimization testing, we provide insights on the neural network architecture, activation functions and optimizers which provide the best results. We show that both Neural ODEs and UDEs can be used effectively for both prediction as well as forecasting of the CWDE ODE system. More importantly, we introduce the "forecasting breakdown point" - the time at which forecasting fails for both Neural ODEs and UDEs. This provides an insight into the applicability of Scientific Machine Learning frameworks in forecasting tasks. 

The paper is structured as follows. We start by presenting the methodology and detailed description for Neural ODEs and UDEs. Subsequently, we present the prediction and forecasting results for the Neural ODEs and UDEs. We also provide hyperparameter optimization plots. Finally, we conclude with a detailed discussion of our results, and the future scope of applying Scientific ML methods in astronomy.

\section{Methodology}
According to Chandrasekhar the equation that governs the structure of degenerate matter in gravitational equilibrium is given by the second-order ordinary differential equation \cite{chandrasekhar1957introduction}
\begin{equation}
    \frac{1}{\eta^2} \frac{d}{d \eta}\left(\eta^2 \frac{d \varphi}{d \eta}\right)+\left(\varphi^2-C\right)^{3 / 2}=0
    \label{WhiteDwarf_ODE}
\end{equation}
with initial conditions
$$
\varphi(0)=1, \quad \varphi^{\prime}(0)=0
$$
This equations is one of Emden type, and therefore a solution exists in the neighborhood of $\eta = 0$ \cite{davis1960introduction}.
This equation exhibits the density $\varphi$ of the White Dwarf as a function of the dimensionless radius $\eta$. Particularly, The variables $\eta$ and $\varphi$ are expected to take real values due to their physical meaning. From this fact, we can entail more restrictions on the behaviour of $\varphi$ and $\eta$ such as their bounds
$$
\sqrt{C}\le \varphi \le 1,
$$
$$
0\le \eta \le \eta_{\infty}
$$
Moreover, the density function is decreasing and tends the lower bound $\sqrt{C}$ i.e 
\begin{equation}
    \lim_{\eta\to\eta_\infty} \varphi(\eta) = \sqrt{C}
\end{equation}

For the computational implementations, the constant $C$ was set to 0.01. The ODE \eqref{WhiteDwarf_ODE}, was reformulated as a system of first order ODEs

\begin{subequations}
\begin{align}
\frac{d\varphi}{d\eta} &= \theta \\
\frac{d\theta}{d\eta} &= -\frac{2}{\eta} \theta - (\varphi^2 - C)^{3/2}
\end{align}
\label{CWDE reformulated}
\end{subequations}

The finite-length $\eta$ interval in which Chandrasekhar's White Dwarf equation is solvable was obtained by implementing a numerical approach in the Julia programming language. For this C value, the set of valid values obtained for $\eta$ was $D_f = \{\eta \in \mathbb{R} : 0.05\le \eta \le 5.325\} =  [0.05, 5.325]$. Subsequently, the domain $D_f$ was discretized into 100 equally spaced $\eta$ values. For these $\eta$ points, the values for both $\varphi$ and $\varphi '$ were saved from the numerical calculation of the ODE, resulting in synthetic data characterizing the White Dwarf for this fixed $C$. 
Additionally, noise was induced into the synthetic data with varying standard deviations, resulting in different training datasets. Specifically, the standard deviations for the added noise were $7\%$ and $35\%$ regarding the synthetic data. These datasets were labeled as moderate-noise data and high-noise data, respectively, while the synthetic data without any added noise was labeled as no-noise data. For the training routines different subsets of these datasets were used to test the forecasting capability of the Neural Network models. Particularly, the training routines were implemented with the entire, $90 \%$, $80 \%$, $40 \%$, $20 \%$, and $10 \%$ of the mentioned datasets.
\subsection{Neural ODEs}
Neural Ordinary Differential Equations (Neural ODEs) are a class of models that represent continuous-depth neural networks. Introduced by \cite{node1}, Neural ODEs have opened up new possibilities in modelling continuous processes by using differential equations to define the evolution of hidden states in neural networks. Neural ODEs represent a novel approach in machine learning and computational modelling that combines neural networks with ordinary differential equations (ODEs). Neural ODEs are a subset of the broader spectrum of Scientific Machine Learning and Physics Informed Machine Learning. The key idea behind Neural ODEs is to use a neural network to approximate the solution of an ODE, thereby allowing for flexible modelling of continuous-time dynamics \cite{nodeapp1, nodeapp2, nodeapp3, nodeapp4, nodeapp5, nodeapp6}.

In a traditional neural network, hidden states are updated using discrete layers. In contrast, Neural ODEs use a continuous transformation defined by an ordinary differential equation:

\begin{equation}\frac{dh}{dt} = \textit{f} (h(t), t , \theta)\end{equation}

where,
\begin{itemize}
    \item \textit{h(t)} is the hidden state at time \textit{t}.
    \item \textit{f} is a neural network parameterized by $\theta$.
    \item The hidden state evolves according to the function \textit{f}.
\end{itemize}

%A set of 100 $\eta$ points was obtained by spliting the interval $D_f$ in 100 equally spaced data points.

%Different datasets were used for the training of the Neural ODEs
One crucial aspect of a Neural ODE model (and machine learning models in general) is hyper-parameter tuning. In this work, the selection of the parameters of the model were obtained after a robust search over a range of possible values specific to the training data used. Regarding the entire no-noise dataset, the selected hyper-parameters for the Neural ODE model are shown in table \ref{tab:NODEs_hyper}
\begin{table}[H]
\begin{center}
\begin{tabular}{lll}
\hline Hyperparameter & Values & Search Range \\
\hline$t_{\text {span }}$ & $(0.05, 5.325)$ & $(0,0.5)-(0,10.0)$ \\
Activation Function &  tanh & ReLU, tanh, sigmoid,  RBF kernel \\
Optimization Solver & Adam, BFGS & Adam, RAdam, BFGS \\
Learning Rate & Adam: 0.1 $\&$ BFGS: 0.01 & $0.01,0.02,0.2,0.05,0.1,0.005, 0.006$\\
Hidden units & 160 & $15, 25,50,100, 160, 240$ \\
Number of Epochs & Adam: 80 $\&$ BFGS: 100 & $50-4000$ \\
Loss & $4.1138406539108517e-4$ & (0,0.2) \\
\hline
\end{tabular}
\caption{Neural ODE range of hyper-parameters on training data (no-noise).  Hyper-parameters for the training routine with the entire available data.}
\label{tab:NODEs_hyper}
\end{center}
\end{table}
\subsection{UDEs}
UDEs (Universal Differential Equations) introduced by \cite{ude1}, combine traditional differential equations with machine learning models, such as neural networks, to create a more flexible and powerful tool for modelling complex systems. This approach integrates the robustness of classical differential equations with the adaptability of neural networks, allowing for more accurate and efficient modelling of systems with unknown or partially known dynamics. UDEs offer improved predictive power by combining data-driven approaches with physical laws \cite{udeapp1, udeapp2, udeapp3, udeapp4}. This is particularly useful in scenarios where purely data-driven models might overfit or fail to generalize. The physical laws embedded in UDEs constrain the learning process, ensuring that the model adheres to known scientific principles. Compared to purely data-driven models, UDEs often require fewer data points to achieve high accuracy. The known differential equations provide a strong prior that guides the learning process, reducing the amount of data needed for training. This efficiency makes UDEs suitable for applications with limited data availability. The UDE model for the Chandrasekhar's White Dwarf equation defined in this work, employed the linear $\theta$ terms in \eqref{CWDE reformulated} as the ground truth model or physical law, as shown in equation \eqref{CWUDE reformulated}

%In the case of the reformulated version of Chandrasekhar's White Dwarf equation \eqref{CWUDE reformulated}, the UDE model used in this work was defined taking some terms of the tra
\begin{subequations}
\begin{align}
\frac{d\varphi}{d\eta} &= \theta + NN_1(P,U)\\
\frac{d\theta}{d\eta} &= -\frac{2}{\eta} \theta + NN_2(P,U)
\end{align}
\label{CWUDE reformulated}
\end{subequations}
Where $P$ are the parameters of the Neural Network (NN) architecture, and $U = \left(\varphi(\eta), \varphi'(\eta)\right)$ are the input parameters.\\ The performance of the trained UDE model can be observed further from the recovered interaction or missing term in the original ODE model \eqref{CWDE reformulated}, i.e $NN1(P_{trained},U)$ and $NN2(P_{trained}, U)$

%\\
%The numerical solution was used to produce a syntetic dataset containing values for $\varphi$ and $\varphi '$ over the domain $(0,\eta_\infty)$ where the ODE equation \eqref{WhiteDwarf_ODE} is solvable. This dataset is labeled as the Ground truth dataset, which in principle should be close to the data measured if we were to run an experiment to obtain the $\varphi$ and $\varphi '$ values for different $\eta$ values in a white dwarf with fixed $C$.

%Introduction
%Deep Neural network models offer a powerful method for modeling phenomena where the analytical expressions are unknown. 
%Neural Network 
%\\ A Neural ODE is a type of deep neural network models. In these models, the derivatives associated with the physical model or phenomena are parameterized using a neural network. This approach offers the advantage of training the associated Neural Network against the dataset obtained either from synthetic or experimental methods. This training shapes the Neural Network parameters to approximate the actual derivatives of the model along with its initial conditions making possible to obtain a Neural Network-based approximation to the actual model. Moreover this Neural Network approximation for solving the ODE was used to forecast $\varphi$ and $\varphi '$  for different unseen $\eta$ values. 

Hyper-parameter tuning is a crucial aspect of the UDE model (and machine learning models in general). In this work, the model's parameters were selected after a robust search over a range of possible values specific to the dataset used for training. Regarding the entire no-noise dataset, the selected hyper-parameters for the UDE model are shown in table \ref{tab:UDEs_hyper}

\begin{table}[H]
\begin{center}
\begin{tabular}{lll}
\hline Hyperparameter & Values & Search Range \\
\hline$t_{\text {span }}$ & $(0.05, 5.325)$ & $(0,0.5)-(0,10.0)$ \\
Activation Function &  RBF kernel & ReLU, tanh, sigmoid,  RBF kernel \\
Optimization Solver & Adam, BFGS & Adam, RAdam, BFGS \\
Learning Rate & Adam: 0.2 $\&$ BFGS: 0.01 & $0.01,0.2,0.001,0.1,0.006, 0.5$ \\
Hidden units & 15 & $15,25,50,100$ \\
Number of Epochs & Adam: 300 $\&$ BFGS: 1000  & $50-4000$ \\
Loss & $7.428771812835665e-8$ & (0,0.2) \\
\hline
\end{tabular}
\caption{UDE range of hyper-parameters on training data (no-noise). Hyper-parameters for the training routine with the entire available data. }
\label{tab:UDEs_hyper}
\end{center}
\end{table}

%Write about the algorithm :

%\label{sec:headings}

%See Section \ref{sec:headings}.
\section{Results}
Six cases were considered for the training process of the deep learning-based models:
\subsection{Case 1: Training in the full domain (100 $\eta$ points)}
\begin{figure}[H]
\centering
\begin{tabular}{cccc}
\includegraphics[width=0.5\textwidth]{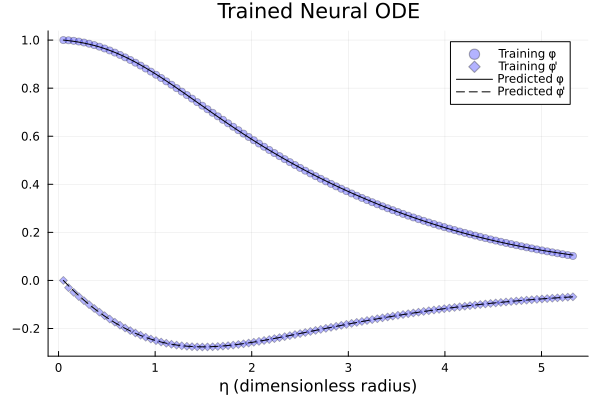} &
\includegraphics[width=0.5\textwidth]{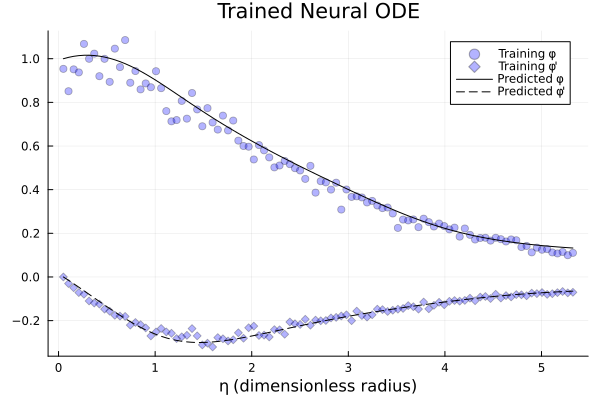} & \\
\text{(a)}  & \text{(b)}  \\[6pt]
\end{tabular}
\begin{tabular}{cccc}
\includegraphics[width=0.5\textwidth]{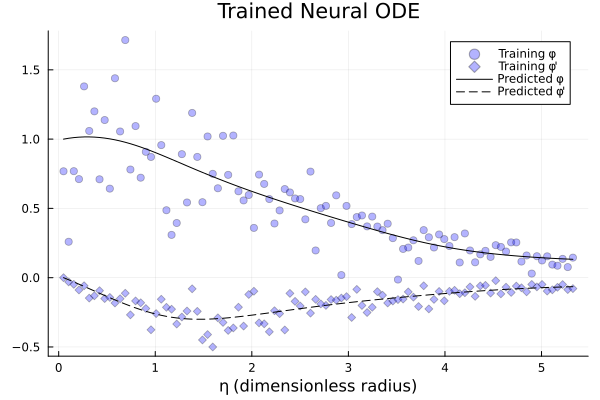} &
 \\
\text{(c)}  \\[6pt]
\end{tabular}
\caption{Comparison of the Neural ODE approximation for the Chandrasekhar’s white
dwarf model. The training of the Neural ODE was performed with varying noise added to the synthetic data in the full solution domain. These training datasets encompassed the values for $\varphi$ and $\varphi'$ at the 100 equally spaced $\eta$ points with varied noise addition. Each figure shows the results for the different training sets:
\text{(a)} No-noise Data (synthetic data) obtained numerically from the White Dwarf ordinary differential equation \eqref{WhiteDwarf_ODE}.
\text{(b)} Moderate-noise dataset with a standard deviation of $7 \%$. 
\text{(c)} High-noise dataset with a standard deviation of $35 \%$. }
\label{NeuralODE_FullDomain}
\end{figure}

\begin{figure}[H]
\centering
\begin{tabular}{cccc}
\includegraphics[width=0.5\textwidth]{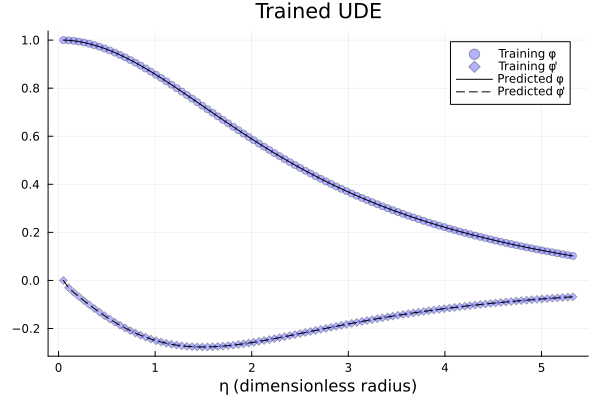} &
\includegraphics[width=0.5\textwidth]{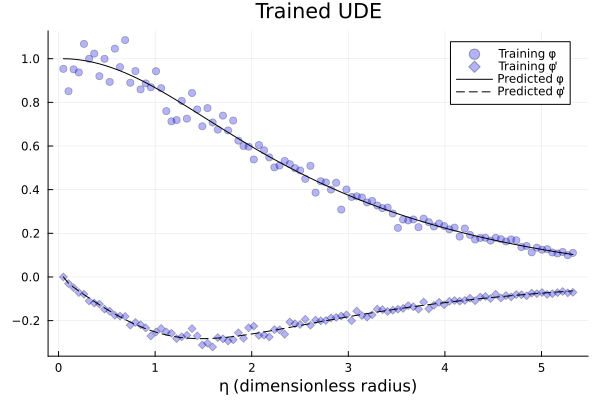} & \\
\text{(a)}  & \text{(b)}  \\[6pt]
\end{tabular}
\begin{tabular}{cccc}
\includegraphics[width=0.5\textwidth]{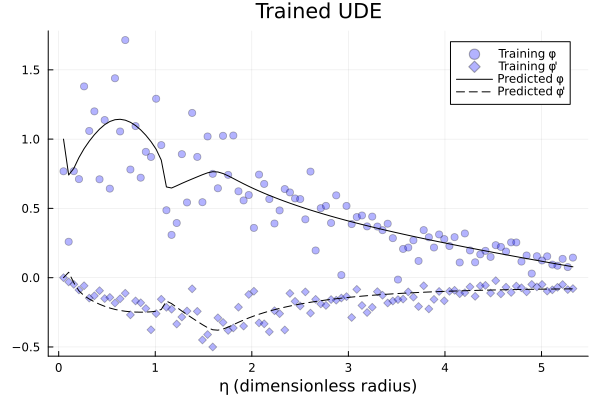} &
 \\
\text{(c)}  \\[6pt]
\end{tabular}
\caption{Comparison of the UDE approximation for the Chandrasekhar's white dwarf equation. The training of the UDE model was performed with varyng noise added to the synthetic data in the full solution domain. These training datasets encompassed the values for $\varphi$ and $\varphi'$ of the 100 equally spaced $\eta$ points with varied noise addition:
\text{(a)} No-noise data (synthetic data) obtained numerically from the White Dwarf ordinary differential equation \eqref{WhiteDwarf_ODE}. 
\text{(b)} Moderate-noise dataset with standard deviation of $7\%$.
\text{(c)} High-noise dataset with standard deviation of $35\%$.}
\label{UDE_FullDomain}
\end{figure}

\begin{figure}[H]
\centering
\begin{tabular}{cccc}
\includegraphics[width=0.5\textwidth]{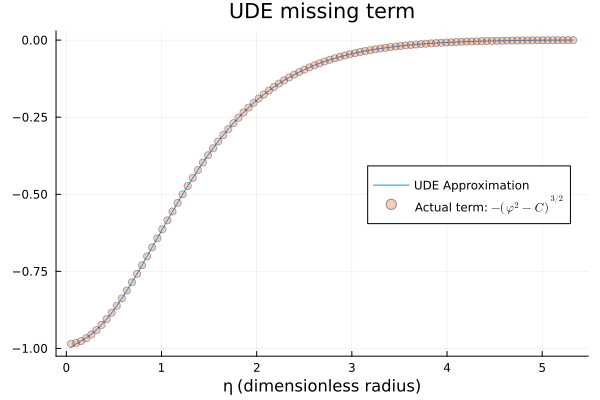} &
\includegraphics[width=0.5\textwidth]{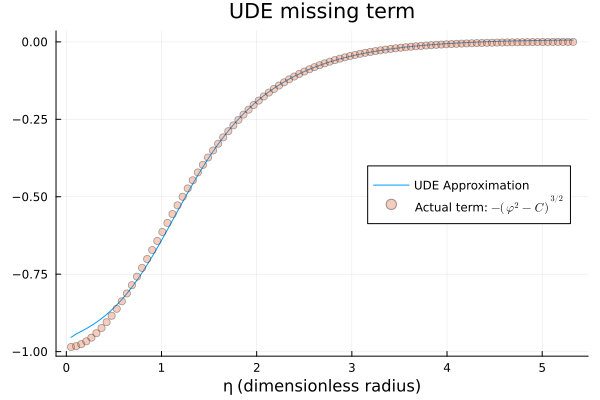} & \\
\text{(a)}  & \text{(b)}  \\[6pt]
\end{tabular}
\begin{tabular}{cccc}
\includegraphics[width=0.5\textwidth]{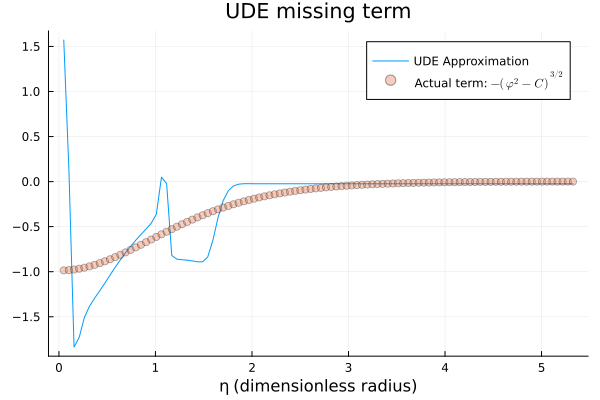} &
 \\
\text{(c)}  \\[6pt]
\end{tabular}
\caption{Comparison of the approximated missing term in the Chandrasekhar's white dwarf UDE model for the different training datasets:
\text{(a)} No-noise dataset (synthetic data) set encompassing the numerically obtained values for $\varphi$ and $\varphi'$ within the solution domain (0, $\eta_\infty$). 
\text{(b)} Moderate-noise dataset with standard deviation of $7\%$ added directly to the synthetic data.
\text{(c)} High-noise dataset with standard deviation of $35\%$ added directly to the synthetic data.}
\label{missingterm_FullDomain}
\end{figure}

First, the implementation of the Neural ODE and UDE models for the Chandrasekhar's White Dwarf equation were performed in the full domain. The three datasets were implemented: no-noise, moderate-noise and high-noise data. The results for the Neural ODE for these training sets are shown in figure \ref{NeuralODE_FullDomain}: \\ \\ \\
We can observe from the graphs in Figure $\ref{NeuralODE_FullDomain}$ that the Neural ODE learns the behavior of Chandrasekhar's White Dwarf equation for both $\varphi$ and $\varphi'$. Even with the addition of moderate and high noise into the dataset, the Neural ODE is still capable of effectively learning the behaviour of the density and its derivative function. However, for the high-noise dataset, the Neural ODE misses the decreasing behaviour of the white dwarf's density function, and it predicts values larger than the initial condition $\varphi (0) = 1$. One distinctive aspect of Chandrasekhar's white Dwarf model is the convergence of the density function $\varphi$ to the square root of the parameter C when $\eta$ approaches the limit where it is defined (5.325 for our training dataset). This convergence is replicated by the Neural ODE approximation for these three datasets.
\\ \\ \\ \\
%---------------------%
%---------------------%
%%UDE For the full Domain figure results
%---------------------%
%---------------------%
\\
The UDE implementation for the Chandrasekhar's model is presented in figure \ref{UDE_FullDomain}
From the last graphics, we can observe that the trained UDE model approximates perfectly the training data for the synthetic set, even with the addition of moderate data (standard deviation of $7\%$), the UDE model can express precisely the behaviour of $\varphi$ and $\varphi'$. For the addition of high-noise in the data (standard deviation of $35\%$), the UDE seems to overfit the training data, leading to a misinterpretation of the $\varphi$ and $\varphi'$ functions for $\eta \in (0,1)$. In spite of this the UDE recovers the converging nature of the density $\varphi$ to the square root of C at the bounding $\eta$ value. \\ 
The performance of the UDE can be observed further from the recovered interaction or missing term in the original ODE model. In this case the missing term happens to be $-(\varphi^2 -C)^{3/2}$. The UDE approximation of this term can be seen in figure \ref{missingterm_FullDomain} for the different training sets \\ \\ \\ \\ \\ \\ \\ \\ \\ \\ \\

%----------------------
%--------------------

%Missing term 
%----------------------
%--------------------
From figure \ref{missingterm_FullDomain}, a perfect match between the approximated term and the actual function $-(\varphi^2 -C)^{3/2}$ can be observed for the no-noise dataset. With the addition of moderate noise ($7 \%$ standard deviation) to the synthetic data, the UDE misses the actual values of the $-(\varphi^2 -C)^{3/2}$ function for the first $7$ $\eta$ points. In the following 3 $\eta$ points the UDE's estimate is slightly off from the actual values but evolves to match almost exactly the actual values for the remaining points, degenerating again for the last 8 $\eta$ points. For the high-noise training dataset, the UDE mostly misses the actual values for $\eta < 2$. After this point, the UDE approximates the actual term values almost exactly. 
\subsection{Case 2: Training with $90\%$ of the full available data and forecasting}
\begin{figure}[H]
\centering
\begin{tabular}{cccc}
\includegraphics[width=0.5\textwidth]{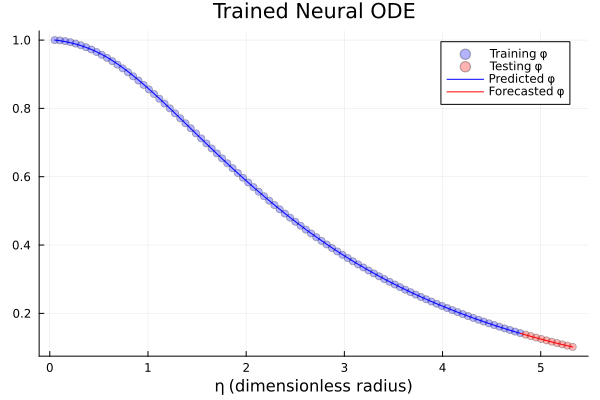} &
\includegraphics[width=0.5\textwidth]{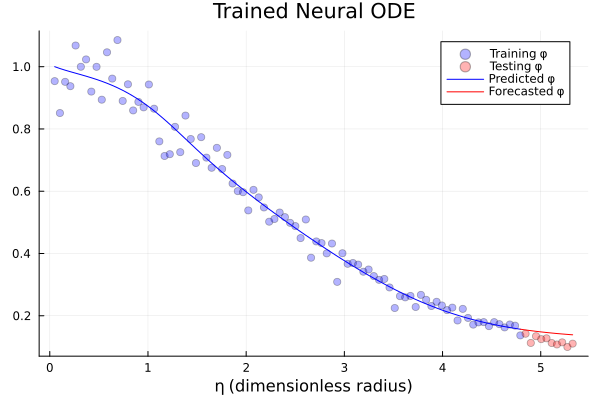} & \\
\text{(a)}  & \text{(b)}  \\[6pt]
\end{tabular}
\begin{tabular}{cccc}
\includegraphics[width=0.5\textwidth]{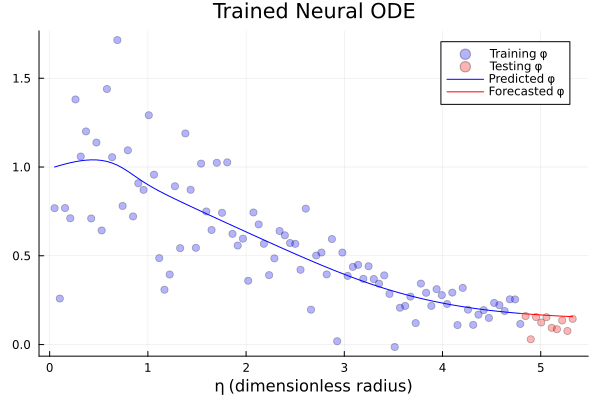} &
 \\
\text{(c)}  \\[6pt]
\end{tabular}
\caption{Comparison of the Neural ODE approximation and forecasting for the Chandrasekhar’s white
dwarf model. The training of the Neural ODE was performed with varying noise added to the synthetic data. These training data subsets encompassed the values for $\varphi$ and $\varphi'$ with varied noise levels added to the first 90 equally spaced $\eta$ points of the solution domain. The forecasted $\varphi$ corresponding to the remaining $10 \%$ of the $\eta$ points are shown against the testing data.
Each figure shows the results for the different datasets:
\text{(a)} No-noise Data (synthetic data) obtained numerically from the White Dwarf ordinary differential equation \eqref{WhiteDwarf_ODE}.
\text{(b)} Moderate-noise dataset with a standard deviation of $7 \%$. 
\text{(c)} High-noise dataset with a standard deviation of $35 \%$.} 
\label{NOde_90Domain}
\end{figure}

 The Neural ODEs and UDEs were trained to evaluate their forecasting capabilities with subsets of the previous no-noise, moderate-noise, and high-noise training datasets. The $\varphi$ and $\varphi'$ values corresponding to the first 90 $\eta$ points from the previous datasets were employed for training. The remaining $\varphi$ and $\varphi'$ values corresponding to the last 10 $\eta$ points were used as testing data to evaluate the UDE forecasting capability in this range. 
For the Neural ODE, the approximated and forecasted density function $\varphi$ for these datasets is shown in figure \ref{NOde_90Domain}\\ \\ \\ \\ \\ \\ 
Figure \ref{NOde_90Domain} shows the forecasting and approximation for the trained Neural ODE model with varying noise concentration datasets. Training with the synthetic data leads to a perfect prediction by the Neural ODE for the $\varphi$ values in the corresponding domain (90 $\%$ of the full $D_f$ interval). Even for the unseen data (the remaining $10 \%$ of the full $D_f$ interval), the Neural ODE forecasts perfectly. For the moderate-noise data, the Neural ODE replicates the behavior of the white dwarf density function and additionally forecasts the unseen data to a good extend. In contrast,  the Neural ODE model performs well with some regions of astray prediction for the high-noise data, especially for $\eta <1 $, where the noise in the training set has a higher magnitude due to its proportion to the $\varphi$ and $\varphi'$ values.

\begin{figure}[H]
\centering
\begin{tabular}{cccc}
\includegraphics[width=0.5\textwidth]{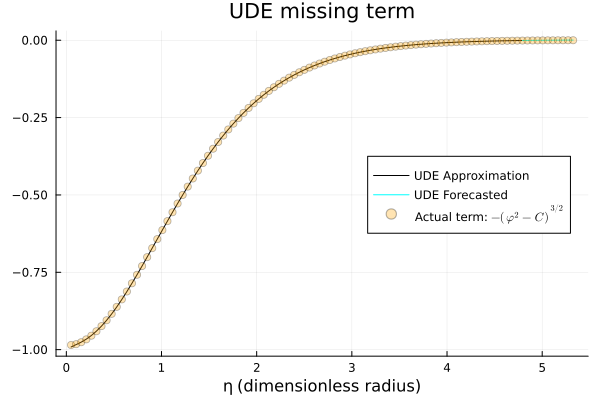} &
\includegraphics[width=0.5\textwidth]{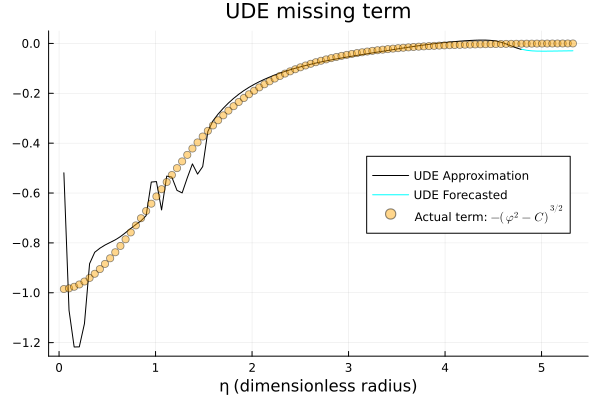} & \\
\text{(a)}  & \text{(b)}  \\[6pt]
\end{tabular}
\begin{tabular}{cccc}
\includegraphics[width=0.5\textwidth]{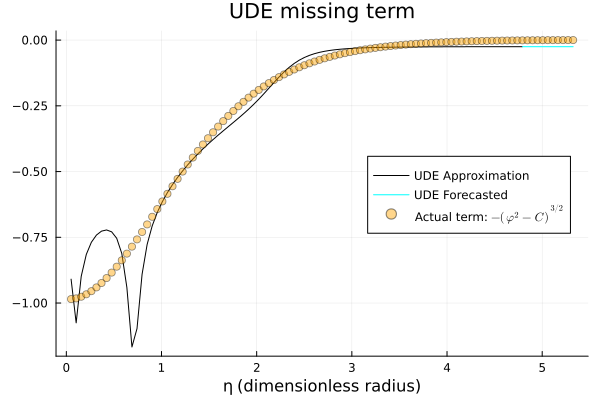} &
 \\
\text{(c)}  \\[6pt]
\end{tabular}
\caption{Comparison of the approximated missing term in the Chandrasekhar's white dwarf UDE model trained with $90 \%$ of the full available datasets:
\text{(a)} No-noise data (synthetic data). 
\text{(b)} Moderate-noise dataset with a standard deviation of $7\%$ added directly to the synthetic data.
\text{(c)} High-noise dataset with a standard deviation of $35\%$ added directly to the synthetic data.}
\label{missingterm_90points}
\end{figure}

% \\
% \newline 
For the trained UDE, the results can be seen in the graphics shown in figure \ref{UDE_90Domain}:

%UDE 90 
\begin{figure}[H]
\centering
\begin{tabular}{cccc}
\includegraphics[width=0.5\textwidth]{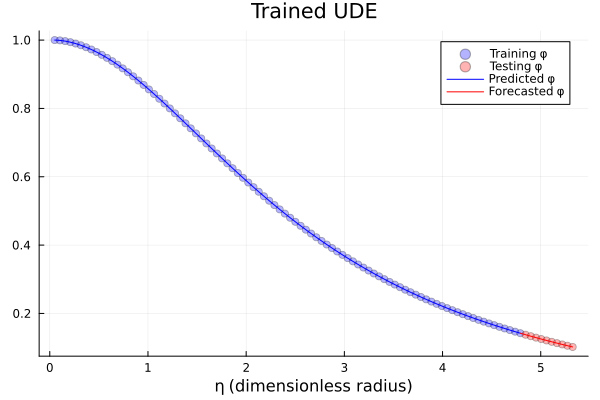} &
\includegraphics[width=0.5\textwidth]{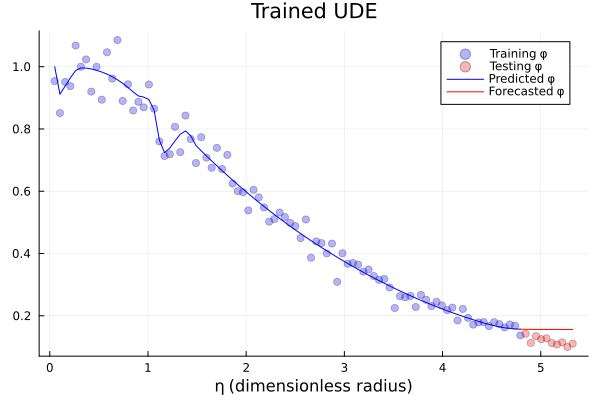} & \\
\text{(a)}  & \text{(b)}  \\[6pt]
\end{tabular}
\begin{tabular}{cccc}
\includegraphics[width=0.5\textwidth]{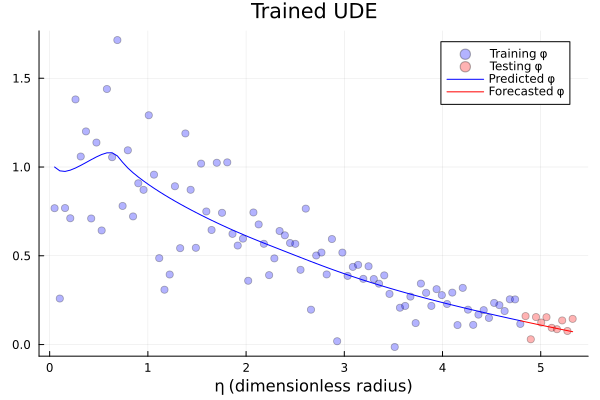} &
 \\
\text{(c)}  \\[6pt]
\end{tabular}
\caption{Comparison of the UDE approximation and forecasting for the Chandrasekhar’s white
dwarf model. The training of the UDE was performed with varying noise added to the synthetic data. These training data subsets encompassed the values for $\varphi$ and $\varphi'$ with varied noise addition for the first 90 equally spaced $\eta$ points of the solution domain. The forecasted $\varphi$ corresponding to the remaining $10 \%$ of the $\eta$ points are shown against the testing data.
Each figure shows the results for the different datasets:
\text{(a)} No-noise data (synthetic data) obtained numerically from the White Dwarf ordinary differential equation \eqref{WhiteDwarf_ODE}.
\text{(b)} Moderate-noise dataset with a standard deviation of $7 \%$. 
\text{(c)} High-noise dataset with a standard deviation of $35 \%$.} 
\label{UDE_90Domain}
\end{figure}
% \\\\\\
The trained UDE performs perfectly, predicting the training data and forecasting the unseen $\varphi$ values for the synthetic dataset. For the moderate-noise data, the UDE reproduces the shape of the actual $\varphi$ function, although some abrupt overfitting due to the noisy data around $\eta = 0$ and $\eta = 1$. 
Finally, the results for the high-noise dataset, shows that the UDE is capable of reproducing the behaviour of the actual values for $\varphi$ despite the noisy data. For the unseen data, it performs similarly to the UDE trained with moderate-noisy data. However, the UDE's performance decreases as the noise level increases. \\
%----------------------
%--------------------
In this training domain, the UDE approximation and forecasting of the missing term are shown in figure \ref{missingterm_90points}
%Missing term 
%----------------------
%--------------------
In figure \ref{missingterm_90points}, the perfect UDE approximation of the missing term in the full domain (both training and testing data) can be seen. This result aligns with the $\varphi$ approximation and forecasting shown in the previous figure (figure \ref{UDE_90Domain}-a). For the moderate-noise data, the approximated missing term is mostly inaccurate for $\eta < 1.2$. However, the UDE accurately approximates the missing term for the remaining $\eta$ values, including the unseen data (cyan). Similarly, for the high-noise dataset, the trained UDE initially approximates the missing term inaccurately but manages to reproduce the actual behavior from $\eta \sim 1$ onward.

%Then the UDE approximates the missing term accurately for the remaining $\eta$ values, even in the case of unseen data (cyan). Similarly, for the high-noise dataset, the trained UDE approximates inaccurately the missing term in the beginning, but manage to reproduce the actual behaviour from $\eta \sim 1$. 
% \\
% \\
% \newline
% \\
% \\
% \\
% \\
% \\

\subsection{Case 3: Training with $80\%$ of the full available data and forecasting}
\begin{figure}[H]
\centering
\begin{tabular}{cccc}
\includegraphics[width=0.5\textwidth]{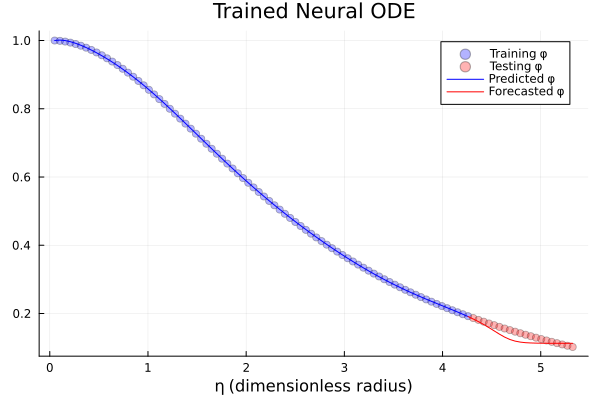} &
\includegraphics[width=0.5\textwidth]{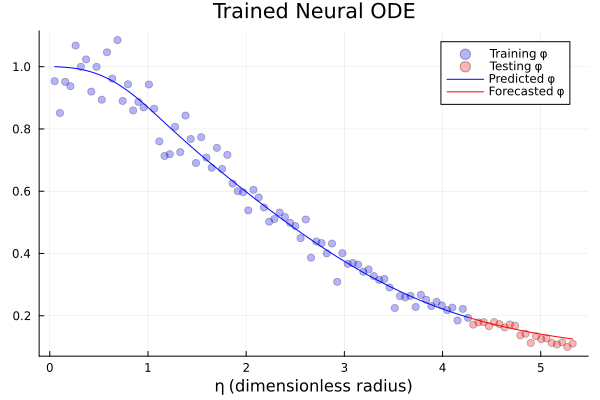} & \\
\text{(a)}  & \text{(b)}  \\[6pt]
\end{tabular}
\begin{tabular}{cccc}
\includegraphics[width=0.5\textwidth]{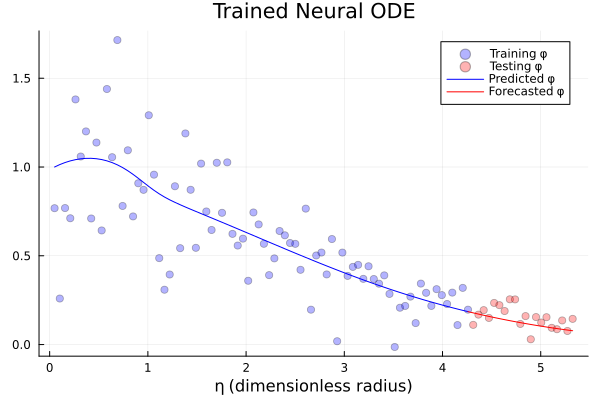} &
 \\
\text{(c)}  \\[6pt]
\end{tabular}
\caption{Comparison of the Neural ODE approximation and forecasting for the Chandrasekhar’s white
dwarf model. The Neural ODE was trained with varying levels of noise added to the synthetic data.  These training data subsets included the values for $\varphi$ and $\varphi'$ with different noise levels for the first 80 equally spaced $\eta$ points of the solution domain. The forecasted $\varphi$ corresponding to the remaining $20 \%$ of the $\eta$ points are shown against the testing data.
Each figure shows the results for the different datasets:
\text{(a)} No-noise data (synthetic data) obtained numerically from the White Dwarf ordinary differential equation \eqref{WhiteDwarf_ODE}.
\text{(b)} Moderate-noise dataset with a standard deviation of $7 \%$. 
\text{(c)} High-noise dataset with a standard deviation of $35 \%$.} 
\label{NeuralODE_80Domain}
\end{figure}

The Neural ODEs and UDEs were trained with smaller data subsets to further evaluate their forecasting capabilities. The previous subsets of no-noise,
moderate-noise, and high-noise training datasets were trimmed, forming new training subsets including the $\varphi$ and $\varphi ' $ values corresponding to the first 80 $\eta$ points of the full domain. The results for the Neural ODE can be seen in thegraphics in figure \ref{NeuralODE_80Domain}.
%Neural ODE results 80 percent of the full data.
The trained Neural ODE approximates the training $\varphi$ data perfectly, but it is slightly off in forecasting the convergence of the density to $\sqrt{C}$.
For the moderate-noise data, the Neural ODE successfully reproduces the behaviour of the $\varphi$ function and its convergence to the square root of C. Finally, for the high-noise data, the Neural ODE manages to recover the shape of the $\varphi$ function and its convergence out of this noisy training data.
\begin{figure}[H]
\centering
\begin{tabular}{cccc}
\includegraphics[width=0.5\textwidth]{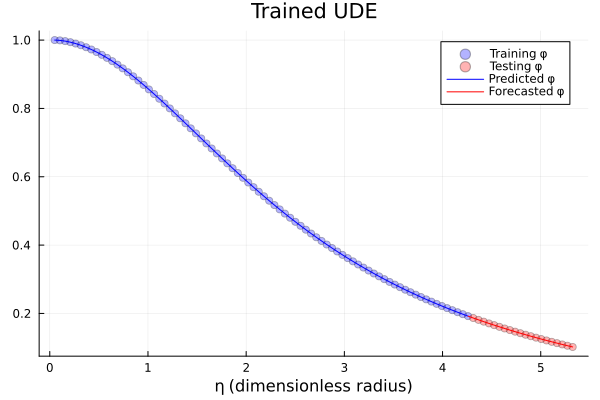} &
\includegraphics[width=0.5\textwidth]{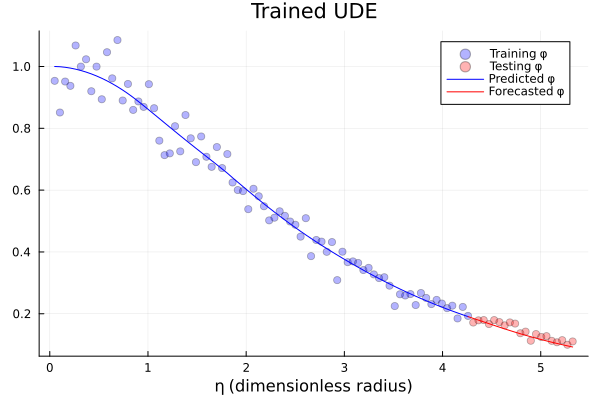} & \\
\text{(a)}  & \text{(b)}  \\[6pt]
\end{tabular}
\begin{tabular}{cccc}
\includegraphics[width=0.5\textwidth]{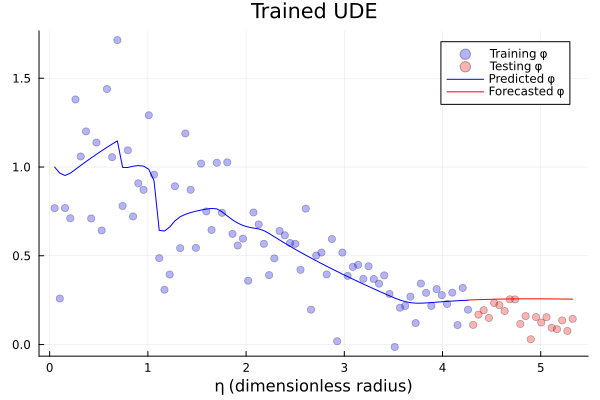} &
 \\
\text{(c)}  \\[6pt]
\end{tabular}
\caption{Comparison of the UDE approximation and forecasting for the Chandrasekhar’s white
dwarf model. The UDE was trained with varying levels of noise added to the synthetic data. These training data subsets included the values for $\varphi$ and $\varphi'$ different noise levels for the first 80 equally spaced $\eta$ points of the solution domain. The forecasted $\varphi$ corresponding to the remaining $20 \%$ of the $\eta$ points are shown against the testing data.
Each figure shows the results for the different datasets:
\text{(a)} No-noise data (synthetic data) obtained numerically from the White Dwarf ordinary differential equation \eqref{WhiteDwarf_ODE}.
\text{(b)} Moderate-noise dataset with a standard deviation of $7 \%$. 
\text{(c)} High-noise dataset with a standard deviation of $35 \%$.} 
\label{UDE_80Domain}
\end{figure}
In figure \ref{UDE_80Domain}, we observe that the UDE accurately approximates the training data and forecasts the unseen data for the no-noise dataset. Similarly, the UDE model performs perfectly for the moderate-noise dataset. However, with the addition of high-noise to the dataset, there is a noticeable breakdown in UDE performance. The UDE tends to overfit abruptly the training data and fails forecasting the unseen values (testing data). 
%----------------------
%--------------------

%Missing term 
%----------------------
%--------------------
\begin{figure}[H]
\centering
\begin{tabular}{cccc}
\includegraphics[width=0.5\textwidth]{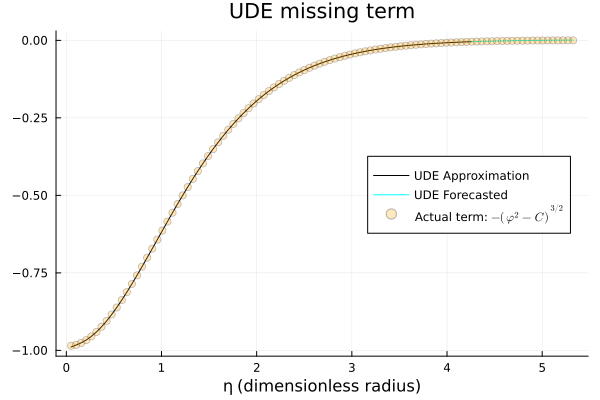} &
\includegraphics[width=0.5\textwidth]{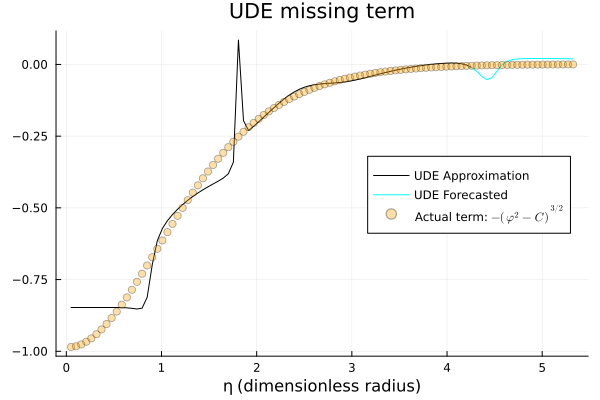} & \\
\text{(a)}  & \text{(b)}  \\[6pt]
\end{tabular}
\begin{tabular}{cccc}
\includegraphics[width=0.5\textwidth]{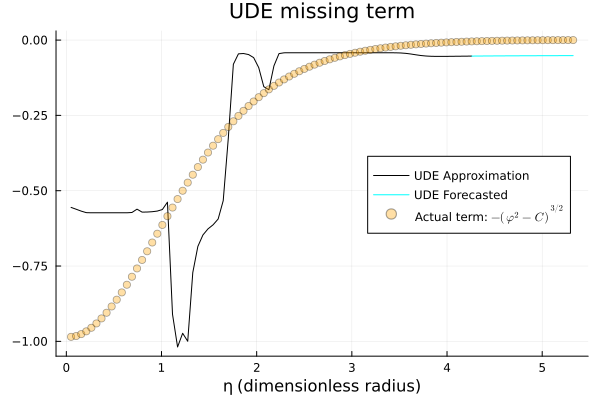} &
 \\
\text{(c)}  \\[6pt]
\end{tabular}
\caption{Comparison of the approximated missing term in the Chandrasekhar's white dwarf UDE model trained with $80 \%$ of the full available datasets:
\text{(a)} No-noise data (synthetic data). 
\text{(b)} Moderate-noise dataset with a standard deviation of $7\%$ added directly to the synthetic data.
\text{(c)} High-noise dataset with a standard deviation of $35\%$ added directly to the synthetic data.}
\label{missingterm_80points}
\end{figure}

In figure \ref{missingterm_80points}, we observe the perfect UDE approximation and forecasting of the missing term for the no-noise dataset. This perfect performance is expected beforehand, as it aligns with the perfect prediction of the $\varphi$ values shown in figure \ref{UDE_80Domain}-a. For the moderate-noise data, the missing terms is mostly well approximated across the full domain $D_f$, although there are some discrepancies. In contrast, the performance for the high-noise data is deficient for most $\eta$ values. This is reflected in the poor approximation and forecasting of $\varphi$ in the previous plot (\ref{UDE_80Domain} -c)  where the results are inaccurate across the full domain. The poor performance with the addition of high noise to the data can be attributed to the overfitting of the model, which causes it to learn the noise rather than the underlying patterns in the data.

%--------------------------------------------
%-----------------------------------------%
%Training with 40 percent of the data 
%--------------------------------------------
%--------------------------------------------
\subsection{Case 4: Training with $40\%$ of the full available data and forecasting}
\begin{figure}[H]
\centering
\begin{tabular}{cccc}
\includegraphics[width=0.5\textwidth]{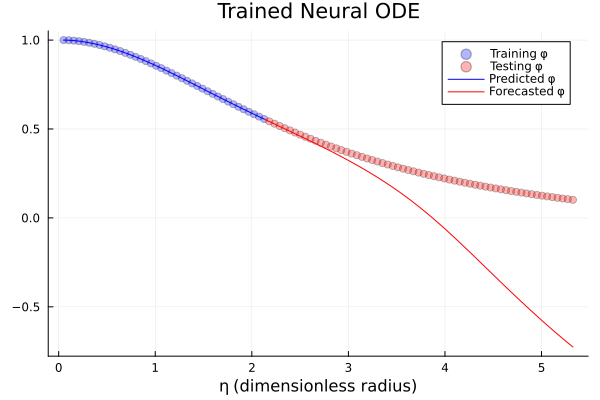} &
\includegraphics[width=0.5\textwidth]{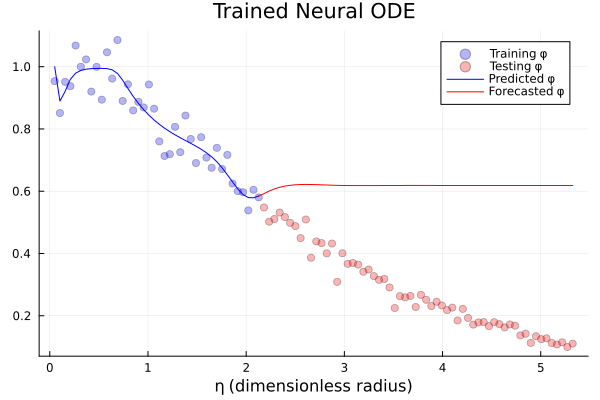} & \\
\text{(a)}  & \text{(b)}  \\[6pt]
\end{tabular}
\begin{tabular}{cccc}
\includegraphics[width=0.5\textwidth]{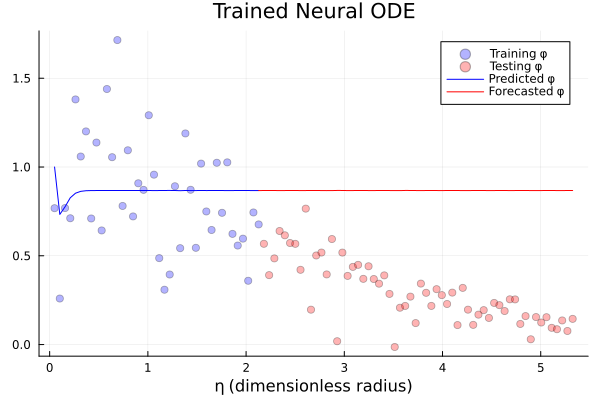} &
 \\
\text{(c)}  \\[6pt]
\end{tabular}
\caption{Comparison of the Neural ODE approximation and forecasting for the Chandrasekhar’s white
dwarf model. The Neural ODE was trained with varying levels of noise added to the synthetic data.  These training data subsets included the values for $\varphi$ and $\varphi'$ with different noise levels for the first 40 equally spaced $\eta$ points of the solution domain. The forecasted $\varphi$ values corresponding to the remaining $60 \%$ of the $\eta$ points are shown against the testing data.
Each figure shows the results for the different datasets:
\text{(a)} No-noise data (synthetic data) obtained numerically from the White Dwarf ordinary differential equation \eqref{WhiteDwarf_ODE}.
\text{(b)} Moderate-noise dataset with a standard deviation of $7 \%$. 
\text{(c)} High-noise dataset with a standard deviation of $35 \%$.} 
\label{NeuralODE_40Domain}
\end{figure}

The Neural ODEs and UDEs were trained with smaller data subsets to further evaluate their forecasting capabilities. The previous subsets of no-noise,
moderate-noise, and high-noise training datasets were trimmed, forming new training data subsets with the $\varphi$ and $\varphi ' $ values corresponding to the first 40 $\eta$ points of the full domain. The results for the Neural ODE can be seen in the graphics in figure \ref{NeuralODE_40Domain}.

In figure \ref{NeuralODE_40Domain}, we observe the breaking point for the Neural ODE model. For all the three datasets, the Neural ODE fails to accurately forecast. In general, the performance worsens with increased noise.  The Neural ODE manages to approximates the training data well for the no-noise dataset, and overfits the moderate-noise training data. whereas for the high-noise data it completely fails. Still for the first two datasets, it misses the forecasting completely. In all these three cases the Neural ODE misses the forcasting of the unseen data significantly.
%%--------------------------------
%%--------------------------------
%Training with 40 percent of the DATa UDE. 
\begin{figure}[H]
\centering
\begin{tabular}{cccc}
\includegraphics[width=0.5\textwidth]{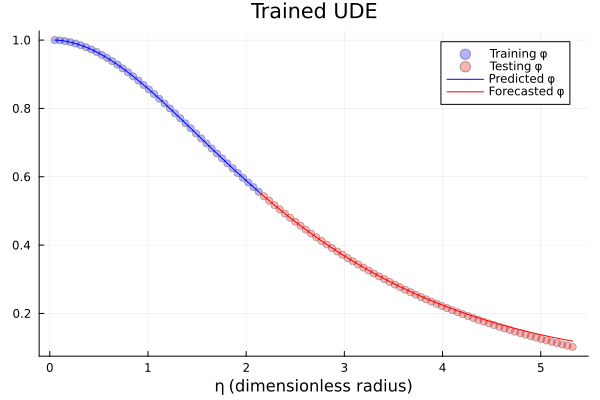} &
\includegraphics[width=0.5\textwidth]{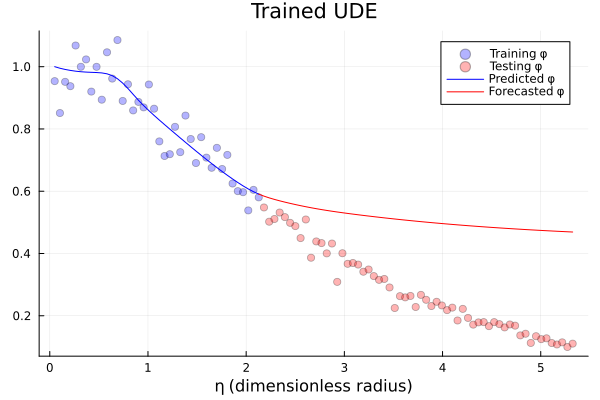} & \\
\text{(a)}  & \text{(b)}  \\[6pt]
\end{tabular}
\begin{tabular}{cccc}
\includegraphics[width=0.5\textwidth]{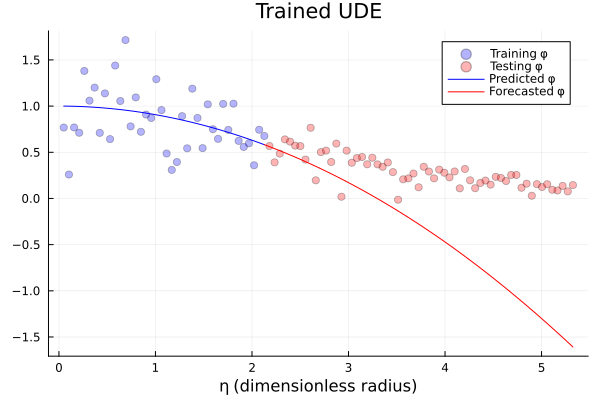} &
 \\
\text{(c)}  \\[6pt]
\end{tabular}
\caption{Comparison of the UDE approximation and forecasting for the Chandrasekhar’s white
dwarf model. The UDE was trained with varying levels of noise added to the synthetic data. These training datasets included the values for $\varphi$ and $\varphi'$ with different noise levels for the first 40 equally spaced $\eta$ points of the solution domain. The forecasted $\varphi$ values corresponding to the remaining $60 \%$ of the $\eta$ points are shown against the testing data.
Each figure shows the results for the different datasets:
\text{(a)} No-noise data (synthetic data) obtained numerically from the White Dwarf ordinary differential equation \eqref{WhiteDwarf_ODE}.
\text{(b)} Moderate-noise dataset with a standard deviation of $7 \%$. 
\text{(c)} High-noise dataset with a standard deviation of $35 \%$.} 
\label{UDE_40Domain}
\end{figure}
From the last figure, we can conclude that the UDE model effectively approximates and forecasts the actual $\varphi$ values when trained with $40 \%$ of the synthetic dataset. However, it fails when noise is added to this fraction of the available data. 

%----------------------
%--------------------

%Missing term 
%----------------------
%--------------------
\begin{figure}[H]
\centering
\begin{tabular}{cccc}
\includegraphics[width=0.5\textwidth]{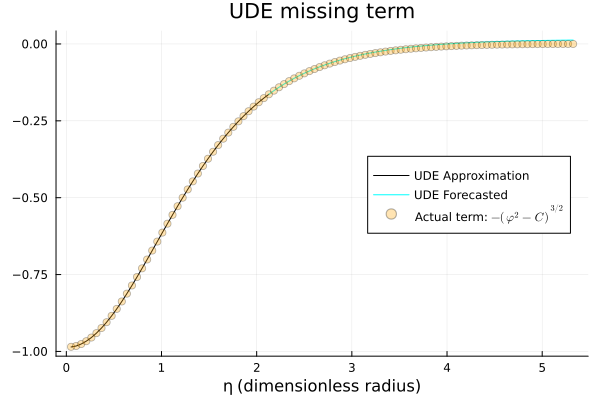} &
\includegraphics[width=0.5\textwidth]{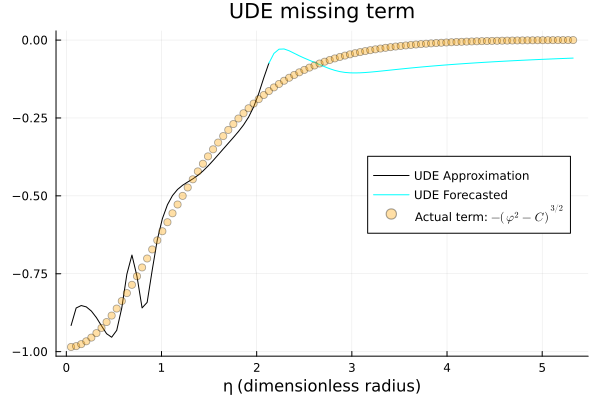} & \\
\text{(a)}  & \text{(b)}  \\[6pt]
\end{tabular}
\begin{tabular}{cccc}
\includegraphics[width=0.5\textwidth]{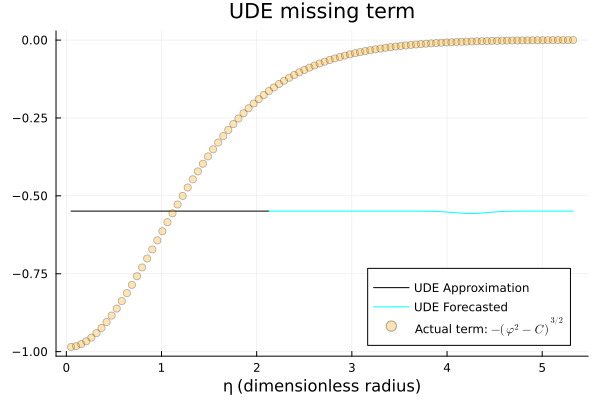} &
 \\
\text{(c)}  \\[6pt]
\end{tabular}
\caption{Comparison of the approximated missing term in the Chandrasekhar's white dwarf UDE model trained with $40 \%$ of the full available datasets:
\text{(a)} No-noise data (synthetic data). 
\text{(b)} Moderate-noise dataset with a standard deviation of $7\%$ added directly to the synthetic data.
\text{(c)} High-noise dataset with a standard deviation of $35\%$ added directly to the synthetic data.}
\label{missingterm_40points}
\end{figure}

From figure \ref{missingterm_40points}, we can conclude several aspects of the UDE model's performance. As expected, for the fraction of the synthetic data used for training, the missing term is properly guessed by the UDE model in the full domain $D_f$. However, with the addition of moderate-noise during the training process, the approximation performance decreases. Moreover, the UDE model’s performance further deteriorates with high noise in the synthetic data.
When comparing the performance of the Neural ODE model to the UDE model, we can see a clear difference. The Neural ODE models fails to forecast unseen data when trained with $40\%$ of each of the three datasets (no-noise data, moderate-noise data, and high-noise data). In contrast, the UDE model performs perfectly when trained with $40\%$ of the total no-noise data, despite its poor performance in forecasting when trained with the same data percentage for the moderate-noise and high-noise datasets. This difference can be attributed to the UDE's encoding of physical laws (the linear terms for the $\varphi '$ variable in \eqref{CWUDE reformulated}), which allows the Neural Network component to address the encoding of the remaining physics from the training data. Consequently, the UDE model forecasts better with less data, especially when the dataset is noise-free (synthetic data). 

%This difference can be attributed to the fact that the UDE encodes physical laws (the linear terms for the $\varphi '$ variable in \eqref{CWUDE reformulated}), producing the Neural Network component to approximated the remaining physics out of the training data, and thus forecast better with less data, especially when the dataset contains no noise (synthetic data). 

%In contrast, the UDE model continues to perform perfectly in the case of training with $40 \%$ of the total no-noise available data, despite its bad performance in forecasting for the moderate-noise data, and the high noise data. The difference can be attributed to the fact that the UDE encodes physics law (the linear terms for the $\varphi '$ variable in \eqref{CWUDE reformulated}), so the UDE guesses the remaining physics and therefore it can forecasts better with less data, specially when no-noise is in the dataset(synthetic data). 
%--------------------------------------------
%-----------------------------------------%
%Training with 20 percent of the data 
%--------------------------------------------
%--------------------------------------------
\subsection{Case 5: Training with $20\%$ of the full available data and forecasting}
\begin{figure}[H]
\centering
\begin{tabular}{cccc}
\includegraphics[width=0.5\textwidth]{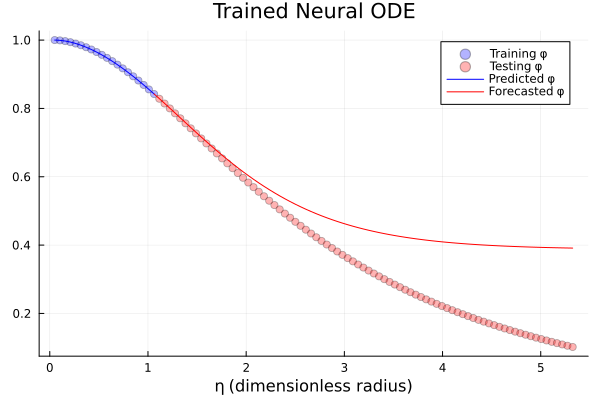} &
\includegraphics[width=0.5\textwidth]{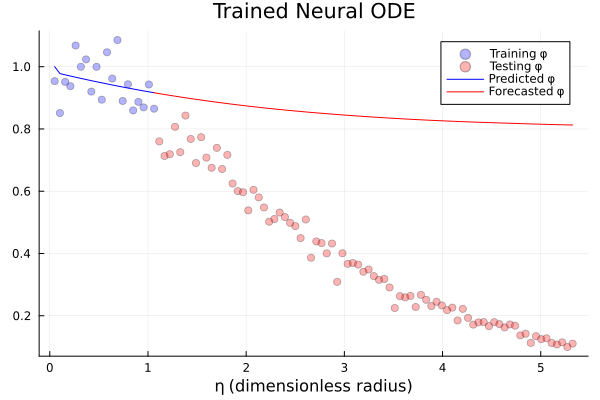} & \\
\text{(a)}  & \text{(b)}  \\[6pt]
\end{tabular}
\begin{tabular}{cccc}
\includegraphics[width=0.5\textwidth]{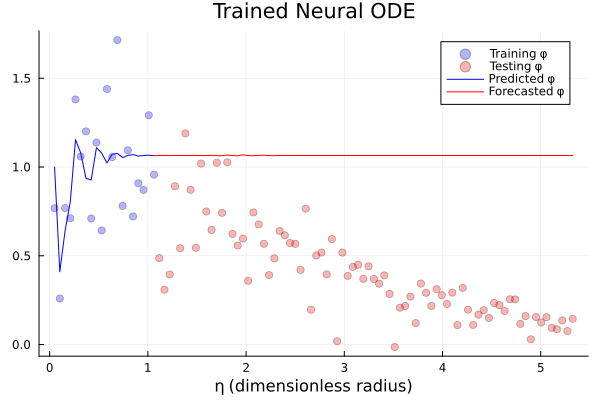} &
 \\
\text{(c)}  \\[6pt]
\end{tabular}
\caption{Comparison of the Neural ODE approximation and forecasting for the Chandrasekhar’s white
dwarf model. The Neural ODE was trained with varying levels of noise added to the synthetic data. These training datasets included the$\varphi$ and $\varphi'$ values  with different noise levels for the first 20 equally spaced $\eta$ points of the solution domain. The forecasted $\varphi$ values for the remaining $80 \%$ of the $\eta$ points are shown against the testing data.
Each figure shows the results for the different datasets:
\text{(a)} No-noise data (synthetic data) obtained numerically from the White Dwarf ordinary differential equation \eqref{WhiteDwarf_ODE}.
\text{(b)} Moderate-noise dataset with a standard deviation of $7 \%$. 
\text{(c)} High-noise dataset with a standard deviation of $35 \%$.} 
\label{NeuralODE_20Domain}
\end{figure}

The Neural ODEs and UDEs were trained with smaller data subsets to further evaluate their forecasting capabilities. The previous subsets of no-noise,
moderate-noise, and high-noise training datasets were trimmed, forming new training data subsets including the $\varphi$ and $\varphi ' $ values corresponding to the first 20 $\eta$ points of the full domain. The results for the Neural ODE can be seen in the graphics in figure \ref{NeuralODE_20Domain}
\\ \\\\ \\ \\ \\ \\
From figure \ref{NeuralODE_20Domain}, we observe the failure of the Neural ODE forecasting the unseen data for the three different datasets. This result is expected, given the already poor performance observed when training with  $40 \%$ of the data. The figure also shows that the performance deteriorates further with the addition of more noise to the synthetic data.

%UDE 20 points
\begin{figure}[H]
\centering
\begin{tabular}{cccc}
\includegraphics[width=0.5\textwidth]{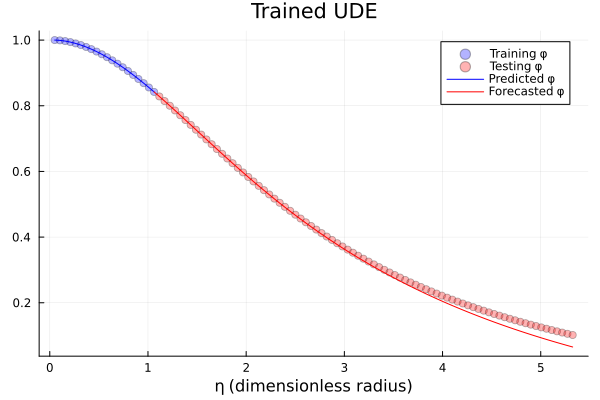} &
\includegraphics[width=0.5\textwidth]{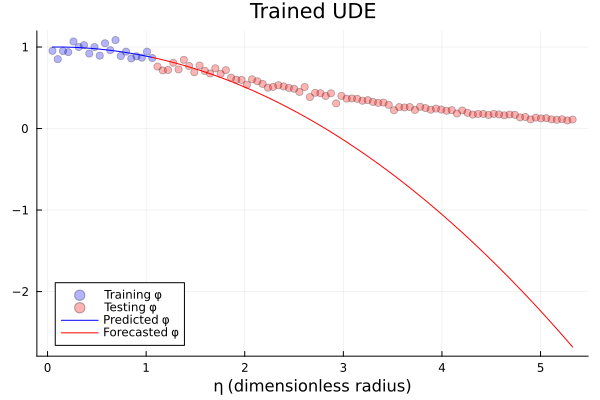} & \\
\text{(a)}  & \text{(b)}  \\[6pt]
\end{tabular}
\begin{tabular}{cccc}
\includegraphics[width=0.5\textwidth]{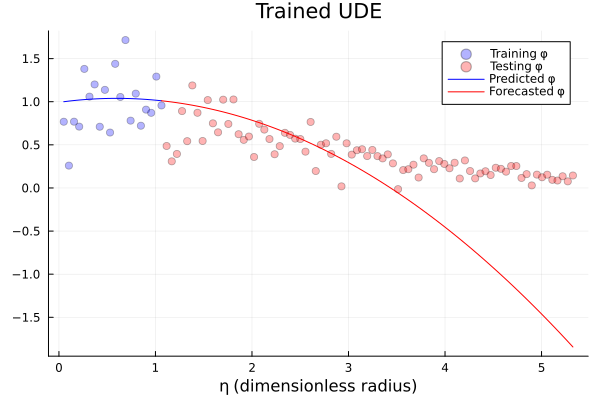} &
 \\
\text{(c)}  \\[6pt]
\end{tabular}
\caption{Comparison of the UDE approximation and forecasting for the Chandrasekhar’s white
dwarf model.The UDE was trained with varying levels of noise added to the synthetic data. These training datasets included the $\varphi$ and $\varphi'$ values with different noise levels for the first 20 equally spaced $\eta$ points of the solution domain. The forecasted $\varphi$ values corresponding to the remaining $80 \%$ of the $\eta$ points are shown against the testing data.
Each figure shows the results for the different datasets:
\text{(a)} No-noise data (synthetic data) obtained numerically from the White Dwarf ordinary differential equation \eqref{WhiteDwarf_ODE}.
\text{(b)} Moderate-noise dataset with a standard deviation of $7 \%$. 
\text{(c)} High-noise dataset with a standard deviation of $35 \%$.} 
\label{UDE_20Domain}
\end{figure}
In figure \ref{UDE_20Domain}, we observe that even with just 20 points, the UDE model successfully approximates and forecasts the actual $\varphi$ values for all $\eta$ in $D_f$. However, the addition of noise (both moderate and high) collapses the UDE performance for the unseen data. 
%----------------------
%--------------------

%Missing term 
%----------------------
%--------------------
\begin{figure}[H]
\centering
\begin{tabular}{cccc}
\includegraphics[width=0.5\textwidth]{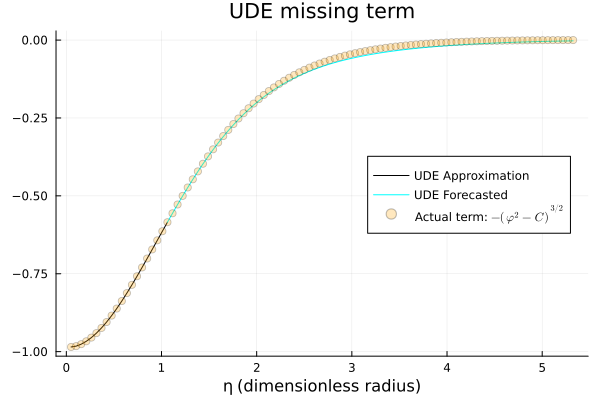} &
\includegraphics[width=0.5\textwidth]{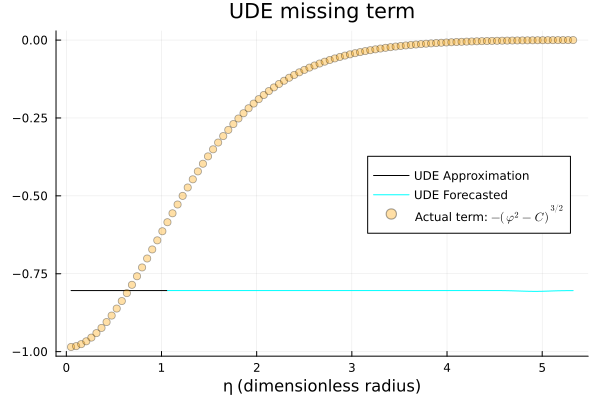} & \\
\text{(a)}  & \text{(b)}  \\[6pt]
\end{tabular}
\begin{tabular}{cccc}
\includegraphics[width=0.5\textwidth]{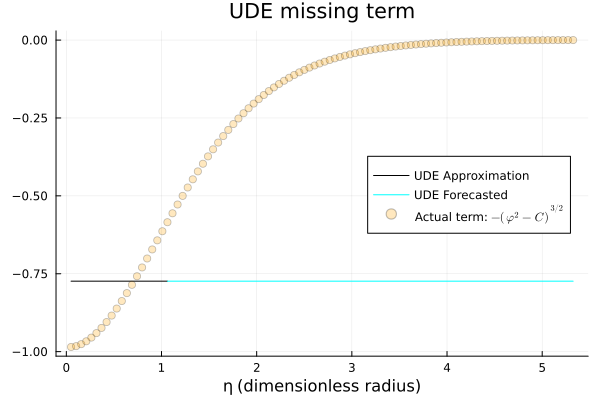} &
 \\
\text{(c)}  \\[6pt]
\end{tabular}
\caption{Comparison of the approximated missing term in the Chandrasekhar's white dwarf UDE model trained with $20 \%$ of the full available datasets:
\text{(a)} No-noise data (synthetic data). 
\text{(b)} Moderate-noise dataset with a standard deviation of $7\%$ added directly to the synthetic data.
\text{(c)} High-noise dataset with a standard deviation of $35\%$ added directly to the synthetic data.}
\label{missingterm_20points}
\end{figure}

For the no-noise training, the UDE excels at approximating and forecasting the values of the missing term in the full domain. However, the addition of noise, completely undermines the UDE's ability to approximate and forecast accurately for all $\eta$ points.

%%%Trainig with 10 percent of the Data
\subsection{Case 6: Training with $10\%$ percent of the full available data and forecasting}
\begin{figure}[H]
\centering
\begin{tabular}{cccc}
\includegraphics[width=0.5\textwidth]{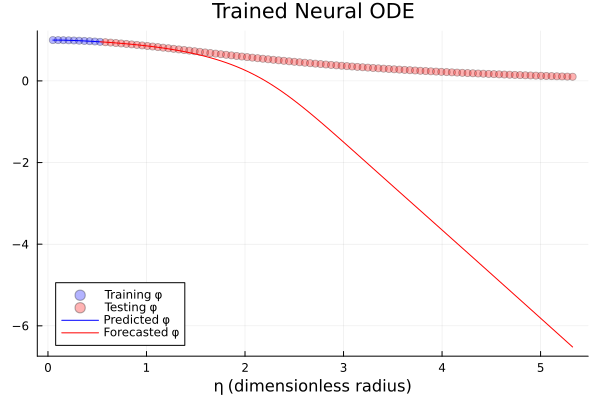} &
\includegraphics[width=0.5\textwidth]{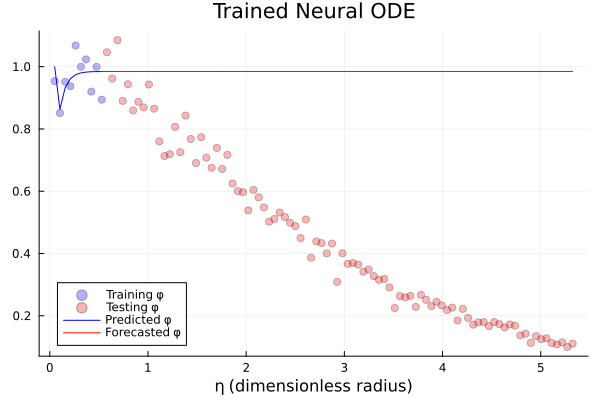} & \\
\text{(a)}  & \text{(b)}  \\[6pt]
\end{tabular}
\begin{tabular}{cccc}
\includegraphics[width=0.5\textwidth]{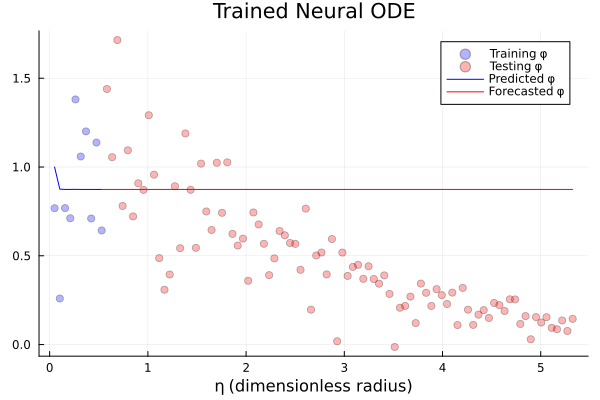} &
 \\
\text{(c)}  \\[6pt]
\end{tabular}
\caption{Comparison of the Neural ODE approximation and forecasting for the Chandrasekhar’s white
dwarf model. The Neural ODE was trained with varying levels of noise added to the synthetic data. These training datasets included the $\varphi$ and $\varphi'$ values with different noise levels for the first 10 equally spaced $\eta$ points of the solution domain. The forecasted $\varphi$ values corresponding to the remaining $90 \%$ of the $\eta$ points are shown against the testing data.
Each figure shows the results for the different datasets:
\text{(a)} No-noise data (synthetic data) obtained numerically from the White Dwarf ordinary differential equation \eqref{WhiteDwarf_ODE}.
\text{(b)} Moderate-noise dataset with a standard deviation of $7 \%$. 
\text{(c)} High-noise dataset with a standard deviation of $35 \%$.} 
\label{NeuralODE_10Domain}
\end{figure}

The Neural ODEs and UDEs were trained with smaller data subsets to further evaluate their forecasting capabilities. The previous subsets of no-noise,
moderate-noise, and high-noise training datasets were trimmed, forming new training data subsets including the $\varphi$ and $\varphi ' $ values corresponding to the first 10 $\eta$ points of the full domain. The results for the Neural ODE can be seen in the graphics in figure \ref{NeuralODE_10Domain}

%Neural ODE 10 percent of the data
%--------------------------------
%-----------------------------------

As expected, further reducing the training data exacerbates the already deteriorated performance of the Neural ODE for the no-noise, moderate-noise, and high-noise datasets. 
%Trained UDE 10 percent 
%-----------------------------------------
%-----------------------------------------
%-----------------------------------------
%-----------------------------------------
\begin{figure}[H]
\centering
\begin{tabular}{cccc}
\includegraphics[width=0.5\textwidth]{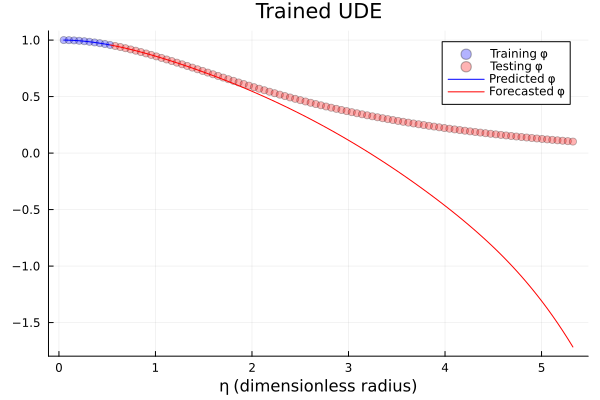} &
\includegraphics[width=0.5\textwidth]{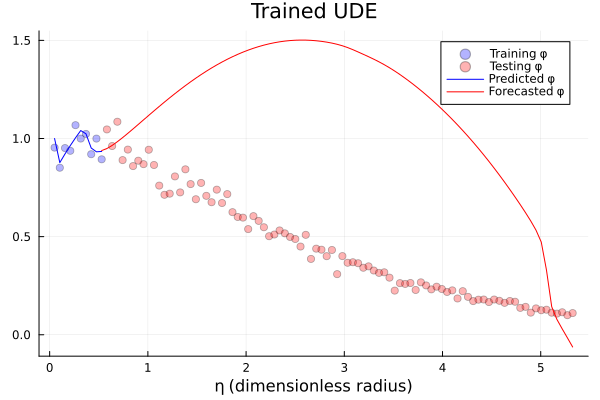} & \\
\text{(a)}  & \text{(b)}  \\[6pt]
\end{tabular}
\begin{tabular}{cccc}
\includegraphics[width=0.5\textwidth]{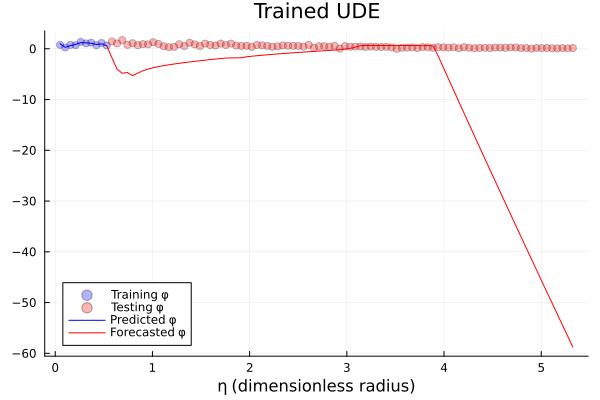} &
 \\
\text{(c)}  \\[6pt]
\end{tabular}
\caption{Comparison of the UDE approximation and forecasting for the Chandrasekhar’s white
dwarf model. The UDE was trained with varying levels of noise added to the synthetic data. These training datasets included the $\varphi$ and $\varphi'$ values with different noise levels for the first 10 equally spaced $\eta$ points of the solution domain. The forecasted $\varphi$ values corresponding to the remaining $90 \%$ of the $\eta$ points are shown against the testing data.
Each figure shows the results for the different datasets:
\text{(a)} No-noise data (synthetic data) obtained numerically from the White Dwarf ordinary differential equation \eqref{WhiteDwarf_ODE}.
\text{(b)} Moderate-noise dataset with a standard deviation of $7 \%$. 
\text{(c)} High-noise dataset with a standard deviation of $35 \%$.} 
\label{UDE_10Domain}
\end{figure}
Reducing the training data to $10 \%$ of the available $\varphi$ and $\varphi'$ values results in a breaking point for the UDE model with no-noise data. The UDE stills manages to forecasts the 20 closest neighboring $\varphi$ values accurately but fails completely beyond this point. For the datasets with added noise, the performance deteriorates further compared to training with a larger number of data points.
\\\\\\

%----------------------
%--------------------

%Missing term 
%----------------------
%--------------------
\begin{figure}[H]
\centering
\begin{tabular}{cccc}
\includegraphics[width=0.5\textwidth]{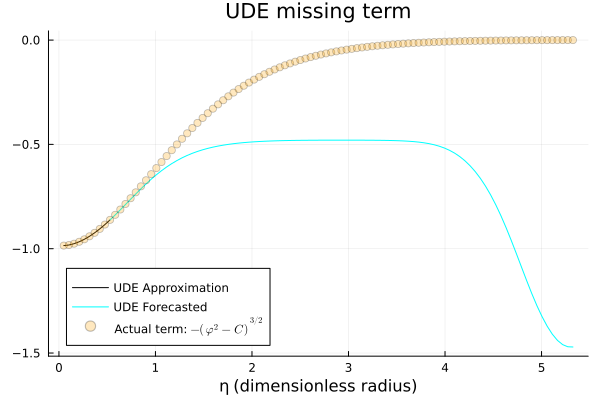} &
\includegraphics[width=0.5\textwidth]{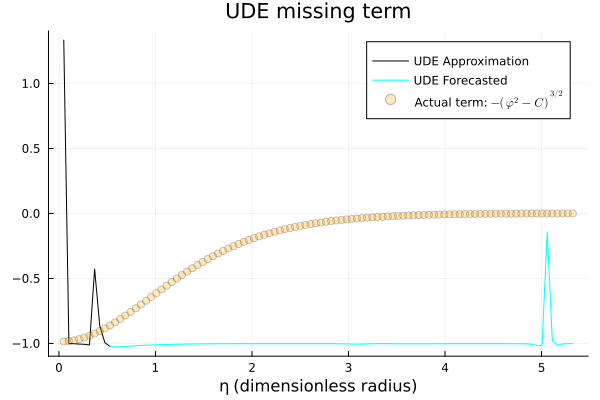} & \\
\text{(a)}  & \text{(b)}  \\[6pt]
\end{tabular}
\begin{tabular}{cccc}
\includegraphics[width=0.5\textwidth]{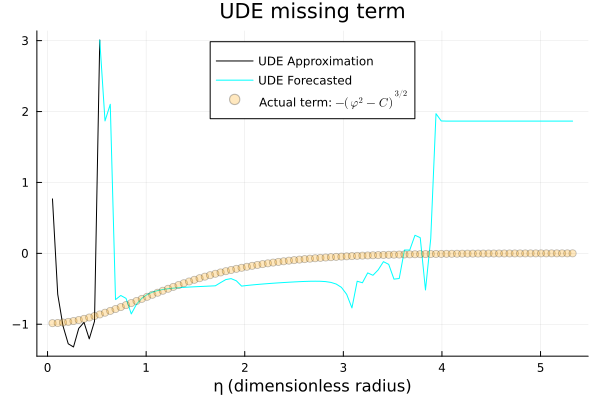} &
 \\
\text{(c)}  \\[6pt]
\end{tabular}
\caption{Comparison of the approximated missing term in the Chandrasekhar's white dwarf UDE model trained with $10 \%$ of the full available datasets:
\text{(a)} No-noise data (synthetic data). 
\text{(b)} Moderate-noise dataset with a standard deviation of $7\%$ added directly to the synthetic data.
\text{(c)} High-noise dataset with a standard deviation of $35\%$ added directly to the synthetic data.}
\label{missingterm_10points}
\end{figure}

For the no-noise training, the UDE accurately predicts the missing term for density values associated with $\eta <1$. However, it misses beyond this $\eta$ point. Similarly, the prediction and forecasting values for the moderate-noise and high-noise training deviate from the missing term's actual values. 
In table \ref{results1} the results for the no-noise fraction of the training datasets are summarized

%Table summarizing the results 
\begin{table}[h]
\centering
\begin{tabular}{lcc}
\toprule
\textbf{Method} & \textbf{Neural ODE} & \textbf{UDE} \\
\midrule
\textbf{Training loss for the full dataset} & 4.1138406539108517e-4 & 7.428771812835665e-8 \\
\textbf{Forecasting breakdown data subset} & $40\%$ of the data & $10\%$ of the data\\
\textbf{Forecasting breakdown $\eta$ point} &  2.128030 & $0.529545$\\
\textbf{Training loss at breakdown point} &  $7.08585918958102e-5$ & $2.7694999610436222e-11$\\
\textbf{Minimum training loss obtained} &  $1.3532552145569704e-5$ & $5.292079273035613e-13$\\
\bottomrule
\end{tabular}
\caption{Summary of performance for the Neural network-based models for the CWDE employed in this work. These results correspond to the null noise fraction of the datasets (synthetic data). }
\label{results1}
\end{table}
In Table \ref{results1}, the first entry corresponds to the loss obtained training with the entire available synthetic dataset. The Forecasting breakdown data subset notes the percentage of the data at which the model broke down. The third entry maps the forecasting breakdown data subset to its corresponding $\eta$ value. The following entry represents the minimum loss obtained when training the models with the breakdown data subsets. The last entry shows the minimum loss obtained across all the training data subsets in the no-noise fraction. The minimum loss for the Neural ODE was obtained when training the model with $10 \%$ of the synthetic data, while for the UDE was when training with $20 \%$ of the data. 
As we can see, the Neural ODE model fails when trained with less than $40\%$ of the available datasets (no-noise, moderate-noise, and high-noise), while the UDE continues to perform well in forecasting even when trained with as little as $20\%$ of the noise-free data, although it fails when noise is added. This superiority in the forecasting capability of the UDE model showcases a deeper understanding of the underlying physical phenomena, which is expected due to the incorporation of physical laws as ground truth for the system. These features make the UDE model more suitable in situations where data availability is limited and part of the mathematical model is known, or when testing trial physical laws in experiments, as it can predict and forecast reliably with less data.

\section{Discussion and Conclusion}
We successfully approximated the underlying data for Chandrasekhar's white dwarf equation (CWDE) with a fixed parameter $C$ using a trained Neural Ordinary Differential Equation (Neural ODE) model with both noiseless and noisy data. A comprehensive study was conducted to identify favorable hyperparameters and neural network architectures. Ultimately, the combination of ADAM and BFGS optimizers, the $\tanh$ activation function, and a streamlined neural network architecture synergistically contributed to a significant improvement of the models' performance (loss reduction). For all datasets (no noise, moderate noise, and high noise), the Neural ODE model effectively approximated the training data. In terms of forecasting, the model performed well in predicting unseen data when trained with at least $80\%$ of the available data. However, the Neural ODE model breaks down and fails to predict the unseen data when trained with less than $40\%$ of the available datasets, indicating that while Neural ODEs offer easier modeling without relying on physical knowledge, they require a substantial amount of data to maintain forecasting reliability.

In contrast, Universal Differential Equations (UDEs) excelled at identifying and recovering the missing term for the CWDE, minimizing the training loss across all datasets.  A comprehensive study was conducted to identify favorable hyperparameters and neural network architectures. Ultimately, the combination of ADAM and BFGS optimizers, the Sigmoid activation function, and a streamlined neural network architecture synergistically contributed to a significant improvement of the models' performance (loss reduction).
UDEs demonstrated superior performance in data-scarce situations, successfully forecasting for all testing values of the dimensionless radius $\eta$ even when trained with just $20\%$ of the noiseless training data. However, similar to the Neural ODE, the UDE model fails to forecast unseen data when trained on less than $40\%$ of the available datasets.

The UDE model's ability to perform well with as little as $20\%$ of noise-free data showcases its efficiency in data-scarce environments, which is particularly valuable in astrophysics where large datasets can be challenging or expensive to obtain. This robustness can be beneficial for simulations in astrophysics due to the UDE efficiency in computational resources. Furthermore, astrophysical data often contain noise from different sources such as instrumental errors, atmospheric disturbances, cosmic rays, and background light. While most of this noise is impossible to avoid, UDE models can be employed in future investigations to model such noise, encoding it using their Neural Network component.  This capability could lead to the identification of noise sources and uncover unknown interactions within the astrophysical system.

%Furthermore, in astrophysical experimental settings, observational data often contain noise from various sources such as instrumental errors, atmospheric disturbances, cosmic rays, and background light, which are impossible to avoid. 
%In this context, UDE models can serve as a tool for modeling such noise by using their neural network component to handle the noise when a verified model of the physical phenomena is known. This could lead to the partial or complete identification of noise sources and even uncover unknown complex interactions within the astrophysical system. 
The UDE model's ability to recover missing interactions or contributions from the training data in this project highlights its potential in the data-driven discovery of missing physics. This capability is crucial in astrophysics, where the exact form of governing equations might not be fully known due to incomplete theories or observational limitations. In this regard, UDEs also provide a valuable tool for theoretical advancements, refining physical laws, and testing hypotheses, therefore offering a new way to investigate the physics of the cosmos.  

%By incorporating neural networks, UDEs can learn and approximate unknown or partially known physics, possibly uncovering unknown interactions, which is particularly useful when dealing with intricate systems like those found in astrophysics.

Beyond their modeling capabilities, The UDE demonstrated strong forecasting power for the CWDE. This suggests that UDEs can enhance the accuracy of predictive models in astrophysics research by leveraging both physical laws (in the form of differential equations) and data-driven corrections (via neural networks). This hybrid approach could improve forecasts of astrophysical events and behaviors, such as stellar evolution, black hole dynamics, or the behavior of neutron stars and white dwarfs. Although the UDE model failed when noise was added in the data-scarce scenario ($40\%$ of the available datasets), its performance with noise-free data suggests that it can be highly effective in controlled experimental settings or simulations, even with low data availability. %

%UDEs also serves as a tool for theoretical advancements, refining physical laws, or testing trial hypotheses more quickly, thereby advancing our understanding of the cosmos.

Applications beyond white dwarfs can be explored, including modeling cosmic ray propagation, understanding galactic dynamics, and simulating accretion processes around black holes. Their versatility makes them a powerful tool for exploring various unsolved problems in astrophysics, offering computational efficiency and effectiveness when data is limited. 

However, UDEs require a well-known physical model to fully leverage their learning power, whereas Neural ODEs are advantageous when no physical law is available, providing a preferred approach in such scenarios.

Overall, the efficiency of UDEs in learning from data and refining model parameters suggests their broad applicability across various scientific and engineering domains, especially those where acquiring data or processing is computationally expensive.

In conclusion, while both Neural ODEs and UDEs effectively capture and predict complex dynamics over shorter intervals, their accuracy and reliability can decline over extended time spans. This underscores the need for continuous model validation and potential adjustments or enhancements to improve long-term forecasting capabilities. The Neural ODE approach offers a black-box-like solution for forecasting and phenomenological modeling when no ground truth is known, but this advantage is contrasted by its need for larger data availability. The UDE model overcomes the problem of extensive data requirement but necessitates a well-known physical model to leverage its learning power. Looking ahead, as scientific machine learning methods are further investigated, a significant focus needs to be placed on forecasting. Most studies in the literature are aimed towards predictions. While the predictive power of SciML methods has been reliably demonstrated, as shown in this study, there are still uncertainties about reliable long-term forecasting. In future work, we will modify the employed SciML models to ensure better forecasting performance, and apply symbolic regression to the recovered terms to discover the symbolic formulations of the terms recovered.
\appendix
\section{Hyperparameters for the training datasets}

\subsection{Case 1: Training with $100 \%$ of the available data.}
\subsubsection*{Neural ODEs}
%No noise data. Neural ODE
\begin{table}[H]
\begin{center}
\begin{tabular}{lll}
\hline Hyperparameter & Values & Search Range \\
\hline$t_{\text {span }}$ & $(0.05, 5.325)$ & $(0,0.5)-(0,10.0)$ \\
Activation Function &  tanh & ReLU, tanh, sigmoid,  RBF kernel \\
Optimization Solver & Adam, BFGS & Adam, RAdam, BFGS \\
Learning Rate & Adam: 0.1 $\&$ BFGS: 0.01 & $0.01,0.02,0.2,0.05,0.1,0.005, 0.006$\\
Hidden units & 160 & $15, 25,50,100, 160, 240$ \\
Number of Epochs & Adam: 80 $\&$ BFGS: 100 & $50-4000$ \\
Loss & $4.1138406539108517e-4$ & (0, 0.2) \\
\hline
\end{tabular}
\caption{Neural ODE range of hyper-parameters on training data (no-noise)}
\label{tab:NODEs_hyper1}
\end{center}
\end{table}
%Moderate noise Neural ODE
\begin{table}[H]
\begin{center}
\begin{tabular}{lll}
\hline Hyperparameter & Values & Search Range \\
\hline$t_{\text {span }}$ & $(0.05, 5.325)$ & $(0,0.5)-(0,10.0)$ \\
Activation Function &  tanh & ReLU, tanh, sigmoid,  RBF kernel \\
Optimization Solver & Adam, BFGS & Adam, RAdam, BFGS \\
Learning Rate & Adam: 0.1 $\&$ BFGS: 0.01 & $0.01,0.02,0.2,0.05,0.1,0.005, 0.006$\\
Hidden units & 160 & $15, 25,50,100, 160, 240$ \\
Number of Epochs & Adam: 80 $\&$ BFGS: 100 & $50-4000$ \\
Loss & 0.18869879897796932  & (0,0.2) \\
\hline
\end{tabular}
\caption{Neural ODE range of hyper-parameters on training data (moderate-noise)}
\label{tab:NODEs_hyper2}
\end{center}
\end{table}

%High noise hyperparameters Neural ODE
\begin{table}[H]
\begin{center}
\begin{tabular}{lll}
\hline Hyperparameter & Values & Search Range \\
\hline$t_{\text {span }}$ & $(0.05, 5.325)$ & $(0,0.5)-(0,10.0)$ \\
Activation Function &  tanh & ReLU, tanh, sigmoid,  RBF kernel \\
Optimization Solver & Adam, BFGS & Adam, RAdam, BFGS \\
Learning Rate & Adam: 0.1 $\&$ BFGS: 0.01 & $0.01,0.02,0.2,0.05,0.1,0.005, 0.006$\\
Hidden units & 160 & $15, 25,50,100, 160, 240$ \\
Number of Epochs & Adam: 80 $\&$ BFGS: 100 & $50-4000$ \\
Loss & 4.6548399612492695 & (0,5.0) \\
\hline
\end{tabular}
\caption{Neural ODE range of hyper-parameters on training data (high-noise)}
\label{tab:NODEs_hyper3}
\end{center}
\end{table}
\subsubsection*{UDEs}
%UDE no noise
\begin{table}[H]
\begin{center}
\begin{tabular}{lll}
\hline Hyperparameter & Values & Search Range \\
\hline$t_{\text {span }}$ & $(0.05, 5.325)$ & $(0,0.5)-(0,10.0)$ \\
Activation Function &  RBF kernel & ReLU, tanh, sigmoid,  RBF kernel \\
Optimization Solver & Adam, BFGS & Adam, RAdam, BFGS \\
Learning Rate & Adam: 0.2 $\&$ BFGS: 0.01 & $0.01,0.2,0.001,0.1,0.006, 0.5$ \\
Hidden units & 15 & $15,25,50,100$ \\
Number of Epochs & Adam: 300 $\&$ BFGS: 1000  & $50-4000$ \\
Loss & $7.428771812835665e-8$ & (0,0.2) \\
\hline
\end{tabular}
\caption{UDE range of hyper-parameters on training data (no-noise)}
\label{tab:UDEs_hyper1}
\end{center}
\end{table}

%moderate noise UDE

\begin{table}[H]
\begin{center}
\begin{tabular}{lll}
\hline Hyperparameter & Values & Search Range \\
\hline$t_{\text {span }}$ & $(0.05, 5.325)$ & $(0,0.5)-(0,10.0)$ \\
Activation Function &  RBF kernel & ReLU, tanh, sigmoid,  RBF kernel \\
Optimization Solver & Adam, BFGS & Adam, RAdam, BFGS \\
Learning Rate & Adam: 0.1 $\&$ BFGS: 0.01 & $0.01,0.2,0.001,0.1,0.006, 0.5$ \\
Hidden units & 15 & $15,25,50,100$ \\
Number of Epochs & Adam: 80 $\&$ BFGS: 100  & $50-4000$ \\
Loss & 0.18700614018769082 & (0,0.2) \\
\hline
\end{tabular}
\caption{UDE range of hyper-parameters on training data (moderate-noise)}
\label{tab:UDEs_hyper2}
\end{center}
\end{table}

%high noise UDE

\begin{table}[H]
\begin{center}
\begin{tabular}{lll}
\hline Hyperparameter & Values & Search Range \\
\hline$t_{\text {span }}$ & $(0.05, 5.325)$ & $(0,0.5)-(0,10.0)$ \\
Activation Function &  RBF kernel & ReLU, tanh, sigmoid,  RBF kernel \\
Optimization Solver & Adam, BFGS & Adam, RAdam, BFGS \\
Learning Rate & Adam: 0.2 $\&$ BFGS: 0.006 & $0.01,0.2,0.001,0.1,0.006, 0.5$ \\
Hidden units & 15 & $15,25,50,100$ \\
Number of Epochs & Adam: 300 $\&$ BFGS: 1000  & $50-4000$ \\
Loss & 3.8823314291961464 & (0,5.0) \\
\hline
\end{tabular}
\caption{UDE Range of hyper-parameters on training data (high-noise)}
\label{tab:UDEs_hyper3}
\end{center}
\end{table}

%case 2

\subsection{Case 2: Training with $90 \%$ of the available data and forecasting.}
%No noise data. Neural ODE
\subsubsection*{Neural ODEs}
\begin{table}[H]
\begin{center}
\begin{tabular}{lll}
\hline Hyperparameter & Values & Search Range \\
\hline$t_{\text {span }}$ & $(0.05, 5.325)$ & $(0,0.5)-(0,10.0)$ \\
Activation Function &  tanh & ReLU, tanh, sigmoid,  RBF kernel \\
Optimization Solver & Adam, BFGS & Adam, RAdam, BFGS \\
Learning Rate & Adam: 0.1 $\&$ BFGS: 0.01 & $0.01,0.02,0.2,0.05,0.1,0.005, 0.006$\\
Hidden units & 160 & $15, 25,50,100, 160, 240$ \\
Number of Epochs & Adam: 80 $\&$ BFGS: 100 & $50-4000$ \\
Loss & 0.0001677152395830407 & (0,0.2) \\
\hline
\end{tabular}
\caption{Neural ODE range of hyper-parameters on training data (no-noise)}
\label{tab:NODEs_hyper4}
\end{center}
\end{table}
%Moderate noise Neural ODE
\begin{table}[H]
\begin{center}
\begin{tabular}{lll}
\hline Hyperparameter & Values & Search Range \\
\hline$t_{\text {span }}$ & $(0.05, 5.325)$ & $(0,0.5)-(0,10.0)$ \\
Activation Function &  tanh & ReLU, tanh, sigmoid,  RBF kernel \\
Optimization Solver & Adam, BFGS & Adam, RAdam, BFGS \\
Learning Rate & Adam: 0.1 $\&$ BFGS: 0.01 & $0.01,0.02,0.2,0.05,0.1,0.005, 0.006$\\
Hidden units & 160 & $15, 25,50,100, 160, 240$ \\
Number of Epochs & Adam: 80 $\&$ BFGS: 100 & $50-4000$ \\
Loss & 0.1922287896870521 & (0,0.2) \\
\hline
\end{tabular}
\caption{Neural ODE range of hyper-parameters on training data (moderate-noise)}
\label{tab:NODEs_hyper5}
\end{center}
\end{table}
%High noise hyperparameters Neural ODE
\begin{table}[H]
\begin{center}
\begin{tabular}{lll}
\hline Hyperparameter & Values & Search Range \\
\hline$t_{\text {span }}$ & $(0.05, 5.325)$ & $(0,0.5)-(0,10.0)$ \\
Activation Function &  tanh & ReLU, tanh, sigmoid,  RBF kernel \\
Optimization Solver & Adam, BFGS & Adam, RAdam, BFGS \\
Learning Rate & Adam: 0.1 $\&$ BFGS: 0.01 & $0.01,0.02,0.2,0.05,0.1,0.005, 0.006$\\
Hidden units & 160 & $15, 25,50,100, 160, 240$ \\
Number of Epochs & Adam: 80 $\&$ BFGS: 100 & $50-4000$ \\
Loss & 4.586295624390106 & (0,5.0) \\
\hline
\end{tabular}
\caption{Neural ODE range of hyper-parameters on training data (high-noise)}
\label{tab:NODEs_hyper6}
\end{center}
\end{table}
\subsubsection*{UDEs}
%UDE no noise
\begin{table}[H]
\begin{center}
\begin{tabular}{lll}
\hline Hyperparameter & Values & Search Range \\
\hline$t_{\text {span }}$ & $(0.05, 5.325)$ & $(0,0.5)-(0,10.0)$ \\
Activation Function &  RBF kernel & ReLU, tanh, sigmoid,  RBF kernel \\
Optimization Solver & Adam, BFGS & Adam, RAdam, BFGS \\
Learning Rate & Adam: 0.2 $\&$ BFGS: 0.01 & $0.01,0.2,0.001,0.1,0.006, 0.5$ \\
Hidden units & 15 & $15,25,50,100$ \\
Number of Epochs & Adam: 300 $\&$ BFGS: 1000  & $50-4000$ \\
Loss & 4.0559441192350175e-8 & (0,0.2) \\
\hline
\end{tabular}
\caption{UDE range of hyper-parameters on training data (no-noise)}
\label{tab:UDEs_hyper4}
\end{center}
\end{table}

%moderate noise UDE

\begin{table}[H]
\begin{center}
\begin{tabular}{lll}
\hline Hyperparameter & Values & Search Range \\
\hline$t_{\text {span }}$ & $(0.05, 5.325)$ & $(0,0.5)-(0,10.0)$ \\
Activation Function &  RBF kernel & ReLU, tanh, sigmoid,  RBF kernel \\
Optimization Solver & Adam, BFGS & Adam, RAdam, BFGS \\
Learning Rate & Adam: 0.2 $\&$ BFGS: 0.01 & $0.01,0.2,0.001,0.1,0.006, 0.5$ \\
Hidden units & 15 & $15,25,50,100$ \\
Number of Epochs & Adam: 300 $\&$ BFGS: 1000  & $50-4000$ \\
Loss & 0.12849386036518934 & (0,0.2) \\
\hline
\end{tabular}
\caption{UDE range of hyper-parameters on training data (moderate-noise)}
\label{tab:UDEs_hyper5}
\end{center}
\end{table}

%high noise UDE

\begin{table}[H]
\begin{center}
\begin{tabular}{lll}
\hline Hyperparameter & Values & Search Range \\
\hline$t_{\text {span }}$ & $(0.05, 5.325)$ & $(0,0.5)-(0,10.0)$ \\
Activation Function &  RBF kernel & ReLU, tanh, sigmoid,  RBF kernel \\
Optimization Solver & Adam, BFGS & Adam, RAdam, BFGS \\
Learning Rate & Adam: 0.2 $\&$ BFGS: 0.01 & $0.01,0.2,0.001,0.1,0.006, 0.5$ \\
Hidden units & 15 & $15,25,50,100$ \\
Number of Epochs & Adam: 300 $\&$ BFGS: 1100  & $50-4000$ \\
Loss & 4.492739863250299 & (0,5.0) \\
\hline
\end{tabular}
\caption{UDE Range of hyper-parameters on training data (high-noise)}
\label{tab:UDEs_hyper6}
\end{center}
\end{table}

%case 3
%
\subsection{Case 3: Training with $80 \%$ of the available data and forecasting.}
%No noise data. Neural ODE
\subsubsection*{Neural ODEs}
\begin{table}[H]
\begin{center}
\begin{tabular}{lll}
\hline Hyperparameter & Values & Search Range \\
\hline$t_{\text {span }}$ & $(0.05, 5.325)$ & $(0,0.5)-(0,10.0)$ \\
Activation Function &  tanh & ReLU, tanh, sigmoid,  RBF kernel \\
Optimization Solver & Adam, BFGS & Adam, RAdam, BFGS \\
Learning Rate & Adam: 0.1 $\&$ BFGS: 0.01 & $0.01,0.02,0.2,0.05,0.1,0.005, 0.006$\\
Hidden units & 160 & $15, 25,50,100, 160, 240$ \\
Number of Epochs & Adam: 80 $\&$ BFGS: 150 & $50-4000$ \\
Loss & 0.0001824757201885788 & (0,0.2) \\
\hline
\end{tabular}
\caption{Neural ODE range of hyper-parameters on training data (no-noise)}
\label{tab:NODEs_hyper7}
\end{center}
\end{table}
%Moderate noise Neural ODE
\begin{table}[H]
\begin{center}
\begin{tabular}{lll}
\hline Hyperparameter & Values & Search Range \\
\hline$t_{\text {span }}$ & $(0.05, 5.325)$ & $(0,0.5)-(0,10.0)$ \\
Activation Function &  tanh & ReLU, tanh, sigmoid,  RBF kernel \\
Optimization Solver & Adam, BFGS & Adam, RAdam, BFGS \\
Learning Rate & Adam: 0.1 $\&$ BFGS: 0.01 & $0.01,0.02,0.2,0.05,0.1,0.005, 0.006$\\
Hidden units & 160 & $15, 25,50,100, 160, 240$ \\
Number of Epochs & Adam: 80 $\&$ BFGS: 100 & $50-4000$ \\
Loss & 0.1814619148898155 & (0,0.2) \\
\hline
\end{tabular}
\caption{Neural ODE range of hyper-parameters on training data (moderate-noise)}
\label{tab:NODEs_hyper8}
\end{center}
\end{table}

%High noise hyperparameters Neural ODE
\begin{table}[H]
\begin{center}
\begin{tabular}{lll}
\hline Hyperparameter & Values & Search Range \\
\hline$t_{\text {span }}$ & $(0.05, 5.325)$ & $(0,0.5)-(0,10.0)$ \\
Activation Function &  tanh & ReLU, tanh, sigmoid,  RBF kernel \\
Optimization Solver & Adam, BFGS & Adam, RAdam, BFGS \\
Learning Rate & Adam: 0.1 $\&$ BFGS: 0.01 & $0.01,0.02,0.2,0.05,0.1,0.005, 0.006$\\
Hidden units & 160 & $15, 25,50,100, 160, 240$ \\
Number of Epochs & Adam: 80 $\&$ BFGS: 150 & $50-4000$ \\
Loss & 4.5409431043449215 & (0,5.0) \\
\hline
\end{tabular}
\caption{Neural ODE range of hyper-parameters on training data (high-noise)}
\label{tab:NODEs_hyper9}
\end{center}
\end{table}
\subsubsection*{UDEs}
%UDE no noise
\begin{table}[H]
\begin{center}
\begin{tabular}{lll}
\hline Hyperparameter & Values & Search Range \\
\hline$t_{\text {span }}$ & $(0.05, 5.325)$ & $(0,0.5)-(0,10.0)$ \\
Activation Function &  RBF kernel & ReLU, tanh, sigmoid,  RBF kernel \\
Optimization Solver & Adam, BFGS & Adam, RAdam, BFGS \\
Learning Rate & Adam: 0.2 $\&$ BFGS: 0.01 & $0.01,0.2,0.001,0.1,0.006, 0.5$ \\
Hidden units & 15 & $15,25,50,100$ \\
Number of Epochs & Adam: 300 $\&$ BFGS: 1000  & $50-4000$ \\
Loss & $6.847620824165587e-9$ & (0,0.2) \\
\hline
\end{tabular}
\caption{UDE range of hyper-parameters on training data (no-noise)}
\label{tab:UDEs_hyper7}
\end{center}
\end{table}

%moderate noise UDE

\begin{table}[H]
\begin{center}
\begin{tabular}{lll}
\hline Hyperparameter & Values & Search Range \\
\hline$t_{\text {span }}$ & $(0.05, 5.325)$ & $(0,0.5)-(0,10.0)$ \\
Activation Function &  RBF kernel & ReLU, tanh, sigmoid,  RBF kernel \\
Optimization Solver & Adam, BFGS & Adam, RAdam, BFGS \\
Learning Rate & Adam: 0.2 $\&$ BFGS: 0.01 & $0.01,0.2,0.001,0.1,0.006, 0.5$ \\
Hidden units & 15 & $15,25,50,100$ \\
Number of Epochs & Adam: 300 $\&$ BFGS: 1300  & $50-4000$ \\
Loss & 0.18109657827904974 & (0,0.2) \\
\hline
\end{tabular}
\caption{UDE range of hyper-parameters on training data (moderate-noise)}
\label{tab:UDEs_hyper8}
\end{center}
\end{table}

%high noise UDE

\begin{table}[H]
\begin{center}
\begin{tabular}{lll}
\hline Hyperparameter & Values & Search Range \\
\hline$t_{\text {span }}$ & $(0.05, 5.325)$ & $(0,0.5)-(0,10.0)$ \\
Activation Function &  RBF kernel & ReLU, tanh, sigmoid,  RBF kernel \\
Optimization Solver & Adam, BFGS & Adam, RAdam, BFGS \\
Learning Rate & Adam: 0.2 $\&$ BFGS: 0.01 & $0.01,0.2,0.001,0.1,0.006, 0.5$ \\
Hidden units & 15 & $15,25,50,100$ \\
Number of Epochs & Adam: 300 $\&$ BFGS: 1100  & $50-4000$ \\
Loss & 3.8463672585184607 & (0,4.0) \\
\hline
\end{tabular}
\caption{UDE range of hyper-parameters on training data (no-noise)}
\label{tab:UDEs_hyper9}
\end{center}
\end{table}

%Case 4 

\subsection{Case 4: Training with $40 \%$ of the available data and forecasting.}
%No noise data. Neural ODE
\subsubsection*{Neural ODEs}
\begin{table}[H]
\begin{center}
\begin{tabular}{lll}
\hline Hyperparameter & Values & Search Range \\
\hline$t_{\text {span }}$ & $(0.05, 5.325)$ & $(0,0.5)-(0,10.0)$ \\
Activation Function &  tanh & ReLU, tanh, sigmoid,  RBF kernel \\
Optimization Solver & Adam, BFGS & Adam, RAdam, BFGS \\
Learning Rate & Adam: 0.02 $\&$ BFGS: 0.01 & $0.01,0.02,0.2,0.05,0.1,0.005, 0.006$\\
Hidden units & 160 & $15, 25,50,100, 160, 240$ \\
Number of Epochs & Adam: 150 $\&$ BFGS: 150 & $50-4000$ \\
Loss & $7.08585918958102e-5$ & (0,0.2) \\
\hline
\end{tabular}
\caption{Neural ODE range of hyper-parameters on training data (no-noise)}
\label{tab:NODEs_hyper10}
\end{center}
\end{table}
%Moderate noise Neural ODE
\begin{table}[H]
\begin{center}
\begin{tabular}{lll}
\hline Hyperparameter & Values & Search Range \\
\hline$t_{\text {span }}$ & $(0.05, 5.325)$ & $(0,0.5)-(0,10.0)$ \\
Activation Function &  tanh & ReLU, tanh, sigmoid,  RBF kernel \\
Optimization Solver & Adam, BFGS & Adam, RAdam, BFGS \\
Learning Rate & Adam: 0.05 $\&$ BFGS: 0.01 & $0.01,0.02,0.2,0.05,0.1,0.005, 0.006$\\
Hidden units & 160 & $15, 25,50,100, 160, 240$ \\
Number of Epochs & Adam: 150 $\&$ BFGS: 300 & $50-4000$ \\
Loss & 0.1168780828919965 & (0,0.2) \\
\hline
\end{tabular}
\caption{Neural ODE range of hyper-parameters on training data (moderate-noise)}
\label{tab:NODEs_hyper11}
\end{center}
\end{table}

%High noise hyperparameters Neural ODE
\begin{table}[H]
\begin{center}
\begin{tabular}{lll}
\hline Hyperparameter & Values & Search Range \\
\hline$t_{\text {span }}$ & $(0.05, 5.325)$ & $(0,0.5)-(0,10.0)$ \\
Activation Function &  tanh & ReLU, tanh, sigmoid,  RBF kernel \\
Optimization Solver & Adam, BFGS & Adam, RAdam, BFGS \\
Learning Rate & Adam: 0.2 $\&$ BFGS: 0.01 & $0.01,0.02,0.2,0.05,0.1,0.005, 0.006$\\
Hidden units & 160 & $15, 25,50,100, 160, 240$ \\
Number of Epochs & Adam: 150 $\&$ BFGS: 300 & $50-4000$ \\
Loss & 4.390035815184278 & (0, 5.0) \\
\hline
\end{tabular}
\caption{Neural ODE range of hyper-parameters on training data (high-noise)}
\label{tab:NODEs_hyper12}
\end{center}
\end{table}
\subsubsection*{UDEs}
%UDE no noise
\begin{table}[H]
\begin{center}
\begin{tabular}{lll}
\hline Hyperparameter & Values & Search Range \\
\hline$t_{\text {span }}$ & $(0.05, 5.325)$ & $(0,0.5)-(0,10.0)$ \\
Activation Function &  RBF kernel & ReLU, tanh, sigmoid,  RBF kernel \\
Optimization Solver & Adam, BFGS & Adam, RAdam, BFGS \\
Learning Rate & Adam: 0.1 $\&$ BFGS: 0.01 & $0.01,0.2,0.001,0.1,0.006, 0.5$ \\
Hidden units & 15 & $15,25,50,100$ \\
Number of Epochs & Adam: 300 $\&$ BFGS: 1000  & $50-4000$ \\
Loss & $3.49521204857231e-10$ & (0,0.2) \\
\hline
\end{tabular}
\caption{UDE range of hyper-parameters on training data (no-noise)}
\label{tab:UDEs_hyper10}
\end{center}
\end{table}

%moderate noise UDE

\begin{table}[H]
\begin{center}
\begin{tabular}{lll}
\hline Hyperparameter & Values & Search Range \\
\hline$t_{\text {span }}$ & $(0.05, 5.325)$ & $(0,0.5)-(0,10.0)$ \\
Activation Function &  RBF kernel & ReLU, tanh, sigmoid,  RBF kernel \\
Optimization Solver & Adam, BFGS & Adam, RAdam, BFGS \\
Learning Rate & Adam: 0.1 $\&$ BFGS: 0.01 & $0.01,0.2,0.001,0.1,0.006, 0.5$ \\
Hidden units & 15 & $15,25,50,100$ \\
Number of Epochs & Adam: 300 $\&$ BFGS: 1500  & $50-4000$ \\
Loss & $0.14576123983603648$ & (0.0.2) \\
\hline
\end{tabular}
\caption{UDE range of hyper-parameters on training data (moderate-noise)}
\label{tab:UDEs_hyper11}
\end{center}
\end{table}

%high noise UDE

\begin{table}[H]
\begin{center}
\begin{tabular}{lll}
\hline Hyperparameter & Values & Search Range \\
\hline$t_{\text {span }}$ & $(0.05, 5.325)$ & $(0,0.5)-(0,10.0)$ \\
Activation Function &  RBF kernel & ReLU, tanh, sigmoid,  RBF kernel \\
Optimization Solver & Adam, BFGS & Adam, RAdam, BFGS \\
Learning Rate & Adam: 0.2 $\&$ BFGS: 0.1 & $0.01,0.2,0.001,0.1,0.006, 0.5$ \\
Hidden units & 15 & $15,25,50,100$ \\
Number of Epochs & Adam: 300 $\&$ BFGS: 200  & $50-4000$ \\
Loss & 3.9559738408639467 & (0,5.0) \\
\hline
\end{tabular}
\caption{UDE range of hyper-parameters on training data (high-noise)}
\label{tab:UDEs_hyper12}
\end{center}
\end{table}
%case 5

\subsection{Case 5: Training with $20 \%$ of the available data and forecasting.}
%No noise data. Neural ODE
\subsubsection*{Neural ODEs}
\begin{table}[H]
\begin{center}
\begin{tabular}{lll}
\hline Hyperparameter & Values & Search Range \\
\hline$t_{\text {span }}$ & $(0.05, 5.325)$ & $(0,0.5)-(0,10.0)$ \\
Activation Function &  tanh & ReLU, tanh, sigmoid,  RBF kernel \\
Optimization Solver & Adam, BFGS & Adam, RAdam, BFGS \\
Learning Rate & Adam: 0.02 $\&$ BFGS: 0.005 & $0.01,0.02,0.2,0.05,0.1,0.005, 0.006$\\
Hidden units & 160 & $15, 25,50,100, 160, 240$ \\
Number of Epochs & Adam: 150 $\&$ BFGS: 125 & $50-4000$ \\
Loss & $6.746089720256949e-5$ & (0,0.2) \\
\hline
\end{tabular}
\caption{Neural ODE range of hyper-parameters on training data (no-noise)}
\label{tab:NODEs_hyper13}
\end{center}
\end{table}
%Moderate noise Neural ODE
\begin{table}[H]
\begin{center}
\begin{tabular}{lll}
\hline Hyperparameter & Values & Search Range \\
\hline$t_{\text {span }}$ & $(0.05, 5.325)$ & $(0,0.5)-(0,10.0)$ \\
Activation Function &  tanh & ReLU, tanh, sigmoid,  RBF kernel \\
Optimization Solver & Adam, BFGS & Adam, RAdam, BFGS \\
Learning Rate & Adam: 0.05 $\&$ BFGS: 0.01 & $0.01,0.02,0.2,0.05,0.1,0.005, 0.006$\\
Hidden units & 160 & $15, 25,50,100, 160, 240$ \\
Number of Epochs & Adam: 150 $\&$ BFGS: 100 & $50-4000$ \\
Loss & 0.09189581738612586 & (0,0.2) \\
\hline
\end{tabular}
\caption{Neural ODE range of hyper-parameters on training data (moderate-noise)}
\label{tab:NODEs_hyper14}
\end{center}
\end{table}

%High noise hyperparameters Neural ODE
\begin{table}[H]
\begin{center}
\begin{tabular}{lll}
\hline Hyperparameter & Values & Search Range \\
\hline$t_{\text {span }}$ & $(0.05, 5.325)$ & $(0,0.5)-(0,10.0)$ \\
Activation Function &  tanh & ReLU, tanh, sigmoid,  RBF kernel \\
Optimization Solver & Adam, BFGS & Adam, RAdam, BFGS \\
Learning Rate & Adam: 0.1 $\&$ BFGS: 0.001 & $0.01,0.02,0.2,0.05,0.1,0.005, 0.006$\\
Hidden units & 160 & $15, 25,50,100, 160, 240$ \\
Number of Epochs & Adam: 100 $\&$ BFGS: 100 & $50-4000$ \\
Loss & 1.4654916015179802 & (0,5.0) \\
\hline
\end{tabular}
\caption{Neural ODE range of hyper-parameters on training data (high-noise)}
\label{tab:NODEs_hyper15}
\end{center}
\end{table}
\subsubsection*{UDEs}
%UDE no noise
\begin{table}[H]
\begin{center}
\begin{tabular}{lll}
\hline Hyperparameter & Values & Search Range \\
\hline$t_{\text {span }}$ & $(0.05, 5.325)$ & $(0,0.5)-(0,10.0)$ \\
Activation Function &  RBF kernel & ReLU, tanh, sigmoid,  RBF kernel \\
Optimization Solver & Adam, BFGS & Adam, RAdam, BFGS \\
Learning Rate & Adam: 0.2 $\&$ BFGS: 0.01 & $0.01,0.2,0.001,0.1,0.006, 0.5$ \\
Hidden units & 15 & $15,25,50,100$ \\
Number of Epochs & Adam: 300 $\&$ BFGS: 1000  & $50-4000$ \\
Loss & $5.292079273035613e-13$ & (0,0.2) \\
\hline
\end{tabular}
\caption{UDE range of hyper-parameters on training data (no-noise)}
\label{tab:UDEs_hyper13}
\end{center}
\end{table}

%moderate noise UDE

\begin{table}[H]
\begin{center}
\begin{tabular}{lll}
\hline Hyperparameter & Values & Search Range \\
\hline$t_{\text {span }}$ & $(0.05, 5.325)$ & $(0,0.5)-(0,10.0)$ \\
Activation Function &  RBF kernel & ReLU, tanh, sigmoid,  RBF kernel \\
Optimization Solver & Adam, BFGS & Adam, RAdam, BFGS \\
Learning Rate & Adam: 0.2 $\&$ BFGS: 0.001 & $0.01,0.2,0.001,0.1,0.006, 0.5$ \\
Hidden units & 15 & $15,25,50,100$ \\
Number of Epochs & Adam: 300 $\&$ BFGS: 1500  & $50-4000$ \\
Loss & 0.08989768088342631 & (0,0.2)\\
\hline
\end{tabular}
\caption{UDE range of hyper-parameters on training data (moderate-noise)}
\label{tab:UDEs_hyper14}
\end{center}
\end{table}

%high noise UDE

\begin{table}[H]
\begin{center}
\begin{tabular}{lll}
\hline Hyperparameter & Values & Search Range \\
\hline$t_{\text {span }}$ & $(0.05, 5.325)$ & $(0,0.5)-(0,10.0)$ \\
Activation Function &  RBF kernel & ReLU, tanh, sigmoid,  RBF kernel \\
Optimization Solver & Adam, BFGS & Adam, RAdam, BFGS \\
Learning Rate & Adam: 0.2 $\&$ BFGS: 0.01 & $0.01,0.2,0.001,0.1,0.006, 0.5$ \\
Hidden units & 15 & $15,25,50,100$ \\
Number of Epochs & Adam: 300 $\&$ BFGS: 1100  & $50-4000$ \\
Loss & 2.1492276515023296 & (0,5.0) \\
\hline
\end{tabular}
\caption{UDE Range of hyper-parameters on training data (high-noise)}
\label{tab:UDEs_hyper15}
\end{center}
\end{table}
%case 6 

\subsection{Case 6: Training with $10 \%$ of the available data and forecasting.}
%No noise data. Neural ODE
\subsubsection*{Neural ODEs}
\begin{table}[H]
\begin{center}
\begin{tabular}{lll}
\hline Hyperparameter & Values & Search Range \\
\hline$t_{\text {span }}$ & $(0.05, 5.325)$ & $(0,0.5)-(0,10.0)$ \\
Activation Function &  tanh & ReLU, tanh, sigmoid,  RBF kernel \\
Optimization Solver & Adam, BFGS & Adam, RAdam, BFGS \\
Learning Rate & Adam: 0.02 $\&$ BFGS: 0.005 & $0.01,0.02,0.2,0.05,0.1,0.005, 0.006$\\
Hidden units & 160 & $15, 25,50,100, 160, 240$ \\
Number of Epochs & Adam: 150 $\&$ BFGS: 125 & $50-4000$ \\
Loss & 1.3532552145569704e-5 & (0,0.2) \\
\hline
\end{tabular}
\caption{Neural ODE range of hyper-parameters on training data (no-noise)}
\label{tab:NODEs_hyper16}
\end{center}
\end{table}
%Moderate noise Neural ODE
\begin{table}[H]
\begin{center}
\begin{tabular}{lll}
\hline Hyperparameter & Values & Search Range \\
\hline$t_{\text {span }}$ & $(0.05, 5.325)$ & $(0,0.5)-(0,10.0)$ \\
Activation Function &  tanh & ReLU, tanh, sigmoid,  RBF kernel \\
Optimization Solver & Adam, BFGS & Adam, RAdam, BFGS \\
Learning Rate & Adam: 0.05 $\&$ BFGS: 0.01 & $0.01,0.02,0.2,0.05,0.1,0.005, 0.006$\\
Hidden units & 160 & $15, 25,50,100, 160, 240$ \\
Number of Epochs & Adam: 150 $\&$ BFGS: 100 & $50-4000$ \\
Loss & 0.03149419050408648 & (0,0.2) \\
\hline
\end{tabular}
\caption{Neural ODE range of hyper-parameters on training data (moderate-noise)}
\label{tab:NODEs_hyper17}
\end{center}
\end{table}

%High noise hyperparameters Neural ODE
\begin{table}[H]
\begin{center}
\begin{tabular}{lll}
\hline Hyperparameter & Values & Search Range \\
\hline$t_{\text {span }}$ & $(0.05, 5.325)$ & $(0,0.5)-(0,10.0)$ \\
Activation Function &  tanh & ReLU, tanh, sigmoid,  RBF kernel \\
Optimization Solver & Adam, BFGS & Adam, RAdam, BFGS \\
Learning Rate & Adam: 0.1 $\&$ BFGS: 0.001 & $0.01,0.02,0.2,0.05,0.1,0.005, 0.006$\\
Hidden units & 160 & $15, 25,50,100, 160, 240$ \\
Number of Epochs & Adam: 100 $\&$ BFGS: 100 & $50-4000$ \\
Loss & 1.035568626897076 & (0,5.0) \\
\hline
\end{tabular}
\caption{Neural ODE range of hyper-parameters on training data (high-noise)}
\label{tab:NODEs_hyper18}
\end{center}
\end{table}
\subsubsection*{UDEs}
%UDE no noise
\begin{table}[H]
\begin{center}
\begin{tabular}{lll}
\hline Hyperparameter & Values & Search Range \\
\hline$t_{\text {span }}$ & $(0.05, 5.325)$ & $(0,0.5)-(0,10.0)$ \\
Activation Function &  RBF kernel & ReLU, tanh, sigmoid,  RBF kernel \\
Optimization Solver & Adam, BFGS & Adam, RAdam, BFGS \\
Learning Rate & Adam: 0.2 $\&$ BFGS: 0.01 & $0.01,0.2,0.001,0.1,0.006, 0.5$ \\
Hidden units & 15 & $15,25,50,100$ \\
Number of Epochs & Adam: 300 $\&$ BFGS: 1000  & $50-4000$ \\
Loss & $2.7694999610436222e-11$ & (0,0.2) \\
\hline
\end{tabular}
\caption{UDE range of hyper-parameters on training data (no-noise)}
\label{tab:UDEs_hyper16}
\end{center}
\end{table}

%moderate noise UDE

\begin{table}[H]
\begin{center}
\begin{tabular}{lll}
\hline Hyperparameter & Values & Search Range \\
\hline$t_{\text {span }}$ & $(0.05, 5.325)$ & $(0,0.5)-(0,10.0)$ \\
Activation Function &  RBF kernel & ReLU, tanh, sigmoid,  RBF kernel \\
Optimization Solver & Adam, BFGS & Adam, RAdam, BFGS \\
Learning Rate & Adam: 0.2 $\&$ BFGS: 0.001 & $0.01,0.2,0.001,0.1,0.006, 0.5$ \\
Hidden units & 15 & $15,25,50,100$ \\
Number of Epochs & Adam: 300 $\&$ BFGS: 1500  & $50-4000$ \\
Loss & 0.017887304919527783 & (0,0.2) \\
\hline
\end{tabular}
\caption{UDE range of hyper-parameters on training data (moderate-noise)}
\label{tab:UDEs_hyper17}
\end{center}
\end{table}

%high noise UDE

\begin{table}[H]
\begin{center}
\begin{tabular}{lll}
\hline Hyperparameter & Values & Search Range \\
\hline$t_{\text {span }}$ & $(0.05, 5.325)$ & $(0,0.5)-(0,10.0)$ \\
Activation Function &  RBF kernel & ReLU, tanh, sigmoid,  RBF kernel \\
Optimization Solver & Adam, BFGS & Adam, RAdam, BFGS \\
Learning Rate & Adam: 0.2 $\&$ BFGS: 0.01 & $0.01,0.2,0.001,0.1,0.006, 0.5$ \\
Hidden units & 15 & $15,25,50,100$ \\
Number of Epochs & Adam: 300 $\&$ BFGS: 200  & $50-4000$ \\
Loss & 0.20057570378189338 & (0,4.0) \\
\hline
\end{tabular}
\caption{UDE range of hyper-parameters on training data (high-noise)}
\label{tab:UDEs_hyper18}
\end{center}
\end{table}

\bibliographystyle{unsrt}
\bibliography{references}

\begin{thebibliography}{10}

\bibitem{scimlappl1}
Nathan Baker, Frank Alexander, Timo Bremer, Aric Hagberg, Yannis Kevrekidis, Habib Najm, Manish Parashar, Abani Patra, James Sethian, Stefan Wild, et~al.
\newblock Workshop report on basic research needs for scientific machine learning: Core technologies for artificial intelligence.
\newblock Technical report, USDOE Office of Science (SC), Washington, DC (United States), 2019.

\bibitem{scimlappl2}
Raj Dandekar, Chris Rackauckas, and George Barbastathis.
\newblock A machine learning-aided global diagnostic and comparative tool to assess effect of quarantine control in covid-19 spread.
\newblock {\em Patterns}, 1(9), 2020.

\bibitem{scimlappl3}
Raj Dandekar, Shane~G Henderson, Marijn Jansen, Sarat Moka, Yoni Nazarathy, Christopher Rackauckas, Peter~G Taylor, and Aapeli Vuorinen.
\newblock Safe blues: A method for estimation and control in the fight against covid-19.
\newblock {\em medRxiv}, pages 2020--05, 2020.

\bibitem{scimlappl4}
Raj Abhijit~Dandekar.
\newblock {\em A new way to do epidemic modeling}.
\newblock PhD thesis, Massachusetts Institute of Technology, 2022.

\bibitem{scimlappl5}
Weiqi Ji, Franz Richter, Michael~J Gollner, and Sili Deng.
\newblock Autonomous kinetic modeling of biomass pyrolysis using chemical reaction neural networks.
\newblock {\em Combustion and Flame}, 240:111992, 2022.

\bibitem{scimlappl6}
Alexander Bills, Shashank Sripad, William~L Fredericks, Matthew Guttenberg, Devin Charles, Evan Frank, and Venkatasubramanian Viswanathan.
\newblock Universal battery performance and degradation model for electric aircraft.
\newblock {\em arXiv preprint arXiv:2008.01527}, 2020.

\bibitem{scimlappl7}
Zhilu Lai, Charilaos Mylonas, Satish Nagarajaiah, and Eleni Chatzi.
\newblock Structural identification with physics-informed neural ordinary differential equations.
\newblock {\em Journal of Sound and Vibration}, 508:116196, 2021.

\bibitem{scimlappl8}
Emily Nieves, Raj Dandekar, and Chris Rackauckas.
\newblock Uncertainty quantified discovery of chemical reaction systems via bayesian scientific machine learning.
\newblock {\em Frontiers in Systems Biology}, 4:1338518, 2024.

\bibitem{scimlappl9}
Allen~M Wang, Darren~T Garnier, and Cristina Rea.
\newblock Hybridizing physics and neural odes for predicting plasma inductance dynamics in tokamak fusion reactors.
\newblock {\em arXiv preprint arXiv:2310.20079}, 2023.

\bibitem{scimlappl10}
Ali Ramadhan.
\newblock {\em Data-driven ocean modeling using neural diferential equations}.
\newblock PhD thesis, Massachusetts Institute of Technology, 2024.

\bibitem{scimlappl11}
Chris~Hill Rackauckas, Jean-Michel Campin, and Raffaele Ferrari.
\newblock Capturing missing physics in climate model.

\bibitem{scimlappl12}
Pushan Sharma, Wai~Tong Chung, Bassem Akoush, and Matthias Ihme.
\newblock A review of physics-informed machine learning in fluid mechanics.
\newblock {\em Energies}, 16(5):2343, 2023.

\bibitem{scimlappl13}
Pushan Sharma, Wai~Tong Chung, Bassem Akoush, and Matthias Ihme.
\newblock A review of physics-informed machine learning in fluid mechanics.
\newblock {\em Energies}, 16(5):2343, 2023.

\bibitem{scimlappl14}
Doaa Aboelyazeed, Chonggang Xu, Forrest~M Hoffman, Jiangtao Liu, Alex~W Jones, Chris Rackauckas, Kathryn Lawson, and Chaopeng Shen.
\newblock A differentiable, physics-informed ecosystem modeling and learning framework for large-scale inverse problems: Demonstration with photosynthesis simulations.
\newblock {\em Biogeosciences}, 20(13):2671--2692, 2023.

\bibitem{node1}
Ricky~TQ Chen, Yulia Rubanova, Jesse Bettencourt, and David~K Duvenaud.
\newblock Neural ordinary differential equations.
\newblock {\em Advances in neural information processing systems}, 31, 2018.

\bibitem{node2}
Emilien Dupont, Arnaud Doucet, and Yee~Whye Teh.
\newblock Augmented neural odes.
\newblock {\em Advances in neural information processing systems}, 32, 2019.

\bibitem{node3}
Stefano Massaroli, Michael Poli, Jinkyoo Park, Atsushi Yamashita, and Hajime Asama.
\newblock Dissecting neural odes.
\newblock {\em Advances in Neural Information Processing Systems}, 33:3952--3963, 2020.

\bibitem{node4}
Hanshu Yan, Jiawei Du, Vincent~YF Tan, and Jiashi Feng.
\newblock On robustness of neural ordinary differential equations.
\newblock {\em arXiv preprint arXiv:1910.05513}, 2019.

\bibitem{ude1}
Christopher Rackauckas, Yingbo Ma, Julius Martensen, Collin Warner, Kirill Zubov, Rohit Supekar, Dominic Skinner, Ali Ramadhan, and Alan Edelman.
\newblock Universal differential equations for scientific machine learning.
\newblock {\em arXiv preprint arXiv:2001.04385}, 2020.

\bibitem{ude2}
Jordi Bolibar, Facundo Sapienza, Fabien Maussion, Redouane Lguensat, Bert Wouters, and Fernando P{\'e}rez.
\newblock Universal differential equations for glacier ice flow modelling.
\newblock {\em Geoscientific Model Development Discussions}, 2023:1--26, 2023.

\bibitem{ude3}
Takeshi Teshima, Koichi Tojo, Masahiro Ikeda, Isao Ishikawa, and Kenta Oono.
\newblock Universal approximation property of neural ordinary differential equations.
\newblock {\em arXiv preprint arXiv:2012.02414}, 2020.

\bibitem{ude4}
Olivier Bournez and Amaury Pouly.
\newblock A universal ordinary differential equation.
\newblock {\em Logical Methods in Computer Science}, 16, 2020.

\bibitem{pinn1}
Maziar Raissi, Paris Perdikaris, and George~E Karniadakis.
\newblock Physics-informed neural networks: A deep learning framework for solving forward and inverse problems involving nonlinear partial differential equations.
\newblock {\em Journal of Computational physics}, 378:686--707, 2019.

\bibitem{pinn2}
Shengze Cai, Zhiping Mao, Zhicheng Wang, Minglang Yin, and George~Em Karniadakis.
\newblock Physics-informed neural networks (pinns) for fluid mechanics: A review.
\newblock {\em Acta Mechanica Sinica}, 37(12):1727--1738, 2021.

\bibitem{pinn3}
George~Em Karniadakis, Ioannis~G Kevrekidis, Lu~Lu, Paris Perdikaris, Sifan Wang, and Liu Yang.
\newblock Physics-informed machine learning.
\newblock {\em Nature Reviews Physics}, 3(6):422--440, 2021.

\bibitem{pinn4}
Shengze Cai, Zhicheng Wang, Sifan Wang, Paris Perdikaris, and George~Em Karniadakis.
\newblock Physics-informed neural networks for heat transfer problems.
\newblock {\em Journal of Heat Transfer}, 143(6):060801, 2021.

\bibitem{astrosciml1}
Raghav Gupta, PK~Srijith, and Shantanu Desai.
\newblock Galaxy morphology classification using neural ordinary differential equations.
\newblock {\em Astronomy and Computing}, 38:100543, 2022.

\bibitem{astrosciml2}
Lorenzo Branca and Andrea Pallottini.
\newblock Neural networks: solving the chemistry of the interstellar medium.
\newblock {\em Monthly Notices of the Royal Astronomical Society}, 518(4):5718--5733, 2023.

\bibitem{astrosciml3}
Sebastien Origer and Dario Izzo.
\newblock Closing the gap: Optimizing guidance and control networks through neural odes.
\newblock {\em arXiv preprint arXiv:2404.16908}, 2024.

\bibitem{cwde1}
Subrahmanyan Chandrasekhar and Subrahmanyan Chandrasekhar.
\newblock {\em An introduction to the study of stellar structure}, volume~2.
\newblock Courier Corporation, 1957.

\bibitem{cwde2}
S~Chandrasekhar.
\newblock The highly collapsed configurations of a stellar mass.
\newblock {\em Neutron Stars, Black Holes and Binary X-Ray Sources}, 48:259, 1975.

\bibitem{Julia1}
Jeff Bezanson, Alan Edelman, Stefan Karpinski, and Viral~B Shah.
\newblock Julia: A fresh approach to numerical computing.
\newblock {\em SIAM review}, 59(1):65--98, 2017.

\bibitem{Julia2}
Jeff Bezanson, Stefan Karpinski, Viral~B Shah, and Alan Edelman.
\newblock Julia: A fast dynamic language for technical computing.
\newblock {\em arXiv preprint arXiv:1209.5145}, 2012.

\bibitem{Julia3}
Kaifeng Gao, Gang Mei, Francesco Piccialli, Salvatore Cuomo, Jingzhi Tu, and Zenan Huo.
\newblock Julia language in machine learning: Algorithms, applications, and open issues.
\newblock {\em Computer Science Review}, 37:100254, 2020.

\bibitem{Julia4}
Chris Rackauckas, Mike Innes, Yingbo Ma, Jesse Bettencourt, Lyndon White, and Vaibhav Dixit.
\newblock Diffeqflux. jl-a julia library for neural differential equations.
\newblock {\em arXiv preprint arXiv:1902.02376}, 2019.

\bibitem{chandrasekhar1957introduction}
Subrahmanyan Chandrasekhar and Subrahmanyan Chandrasekhar.
\newblock {\em An introduction to the study of stellar structure}, volume~2.
\newblock Courier Corporation, 1957.

\bibitem{davis1960introduction}
Harold~Thayer Davis.
\newblock {\em Introduction to nonlinear differential and integral equations}.
\newblock Dover Publications, INC., New york, 1960.

\bibitem{nodeapp1}
Kookjin Lee and Eric~J Parish.
\newblock Parameterized neural ordinary differential equations: Applications to computational physics problems.
\newblock {\em Proceedings of the Royal Society A}, 477(2253):20210162, 2021.

\bibitem{nodeapp2}
Farshud Sorourifar, You Peng, Ivan Castillo, Linh Bui, Juan Venegas, and Joel~A Paulson.
\newblock Physics-enhanced neural ordinary differential equations: Application to industrial chemical reaction systems.
\newblock {\em Industrial \& Engineering Chemistry Research}, 62(38):15563--15577, 2023.

\bibitem{nodeapp3}
Suyong Kim, Weiqi Ji, Sili Deng, Yingbo Ma, and Christopher Rackauckas.
\newblock Stiff neural ordinary differential equations.
\newblock {\em Chaos: An Interdisciplinary Journal of Nonlinear Science}, 31(9), 2021.

\bibitem{nodeapp4}
Mostafa Kiani~Shahvandi, Matthias Schartner, and Benedikt Soja.
\newblock Neural ode differential learning and its application in polar motion prediction.
\newblock {\em Journal of Geophysical Research: Solid Earth}, 127(11):e2022JB024775, 2022.

\bibitem{nodeapp5}
Chris Finlay, J{\"o}rn-Henrik Jacobsen, Levon Nurbekyan, and Adam~M Oberman.
\newblock How to train your neural ode.
\newblock {\em arXiv preprint arXiv:2002.02798}, 2, 2020.

\bibitem{nodeapp6}
Gavin~D Portwood, Peetak~P Mitra, Mateus~Dias Ribeiro, Tan~Minh Nguyen, Balasubramanya~T Nadiga, Juan~A Saenz, Michael Chertkov, Animesh Garg, Anima Anandkumar, Andreas Dengel, et~al.
\newblock Turbulence forecasting via neural ode.
\newblock {\em arXiv preprint arXiv:1911.05180}, 2019.

\bibitem{udeapp1}
Lu~Lu, Pengzhan Jin, and George~Em Karniadakis.
\newblock Deeponet: Learning nonlinear operators for identifying differential equations based on the universal approximation theorem of operators.
\newblock {\em arXiv preprint arXiv:1910.03193}, 2019.

\bibitem{udeapp2}
Mike Innes, Alan Edelman, Keno Fischer, Chris Rackauckas, Elliot Saba, Viral~B Shah, and Will Tebbutt.
\newblock A differentiable programming system to bridge machine learning and scientific computing.
\newblock {\em arXiv preprint arXiv:1907.07587}, 2019.

\bibitem{udeapp3}
Chaopeng Shen, Alison~P Appling, Pierre Gentine, Toshiyuki Bandai, Hoshin Gupta, Alexandre Tartakovsky, Marco Baity-Jesi, Fabrizio Fenicia, Daniel Kifer, Li~Li, et~al.
\newblock Differentiable modelling to unify machine learning and physical models for geosciences.
\newblock {\em Nature Reviews Earth \& Environment}, 4(8):552--567, 2023.

\bibitem{udeapp4}
Chris Rackauckas, Yingbo Ma, Andreas Noack, Vaibhav Dixit, Patrick~Kofod Mogensen, Simon Byrne, Shubham Maddhashiya, Jos{\'e}~Bayo{\'a}n Santiago~Calder{\'o}n, Joakim Nyberg, Jogarao~VS Gobburu, et~al.
\newblock Accelerated predictive healthcare analytics with pumas, a high performance pharmaceutical modeling and simulation platform.
\newblock {\em BioRxiv}, pages 2020--11, 2020.

\end{thebibliography}

\end{document}